\documentclass[mnsc,sglanonrev]{informs4}
\RequirePackage{tgtermes}
\RequirePackage{newtxtext}
\RequirePackage{newtxmath}
\RequirePackage{bm}
\RequirePackage{endnotes}

\OneAndAHalfSpacedXII % Current default line spacing
\usepackage{algorithm}
\usepackage{algpseudocode}
\usepackage{tikz}
\usepackage{booktabs}
\usepackage{algorithm}
\usepackage{algpseudocode}
\usepackage{hyperref}
\usepackage{xcolor}

\usepackage{natbib}
\bibpunct[, ]{(}{)}{,}{a}{}{,}%
\def\bibfont{\small}%
\EquationsNumberedThrough    % Default: (1), (2), ...
\TheoremsNumberedThrough     % Preferred (Theorem 1, Lemma 1, Theorem 2)
\ECRepeatTheorems  %  

\MANUSCRIPTNO{MNSC-0001-2024.00}

\begin{document}
%%%%%%%%%%%%%%%%

%\setcounter{page}{1}
%\DOI{10.1287/xxxx.0000.0000}%

\RUNAUTHOR{Shi et al.}

% \RUNTITLE{Predictive Maintenance in Manufacturing}
%\RUNTITLE{Symphony: A Decentralized Multi-Agent Reinforcement Learning Framework for Market Manipulation Detection via GRPO and Distributed Discourse Reasoning}
\RUNTITLE{Hide-and-Shill: Real-Time Manipulation Detection via MARL in DeFi}

% \TITLE{Optimal Resource Allocation in Humanitarian Logistics: A Stochastic Programming Approach}
%\TITLE{Hide-and-Shill: A Multi-Agent Simulation Framework for Trust-Aware Manipulation Detection Tuned with GRPO}
%\TITLE{Hide-and-Shill: A Reinforcement Learning Framework for Market Manipulation Detection in Symphony—a Decentralized Multi-Agent System}
\TITLE{Hide-and-Shill: A Reinforcement Learning Framework for Market Manipulation Detection in Symphony—a Decentralized Multi-Agent System}

\ARTICLEAUTHORS{%
	\AUTHOR{
		Ronghua Shi\textsuperscript{\textbf{a,$\dagger$}},
		Yiou Liu\textsuperscript{\textbf{b,$\dagger$}},
		Yuchun Feng\textsuperscript{\textbf{e}},
		Lynn Ai\textsuperscript{\textbf{f}},
		Bill Shi\textsuperscript{\textbf{g,*}},
		Zhuang Liu\textsuperscript{\textbf{h,*}},
	}
	\AFF{
		\textsuperscript{\textbf{a}}Department of Information Systems,
		City University of Hong Kong, Hong Kong SAR, China
	}
	\AFF{
		\textsuperscript{\textbf{b}}Business School, University of New South Wales, Sydney, Australia
	}
	\AFF{
		\textsuperscript{\textbf{e}}Division of Engineering Science, University of Toronto, Toronto, Canada
	}
	\AFF{
		\textsuperscript{\textbf{h}}ProphetAI Data Technology Co., Ltd., Beijing, China
	}
	\AFF{
		\textsuperscript{\textbf{f,g}}Gradient, 3 FRASER STREET DUO TOWER, SINGAPORE 
	}
	\textsuperscript{\textbf{*}}Corresponding author. \enspace
	\textsuperscript{$\dagger$}These authors contributed equally to this work.\\
}
\footnotetext[1]{Corresponding author: \textbf{Bill Shi}. Email: \texttt{tianyu@gradient.network}}

% end of the block
\ABSTRACT{%
	Decentralized finance (DeFi) has ushered in a new era of permissionless financial innovation—but also opened the door to discourse-driven market manipulation at unprecedented scale. Without centralized gatekeepers or regulatory oversight, malicious actors now coordinate shilling campaigns and pump-and-dump schemes across social platforms and on-chain ecosystems. We propose \textbf{Hide-and-Shill}, a novel Multi-Agent Reinforcement Learning (MARL) framework for decentralized manipulation detection. By modeling the interaction between manipulators and detectors as a dynamic adversarial game, the framework learns to identify suspicious discourse patterns using delayed token price reactions as ground-truth financial signals.
	Our method introduces three key innovations: (1) \textbf{Group Relative Policy Optimization (GRPO)} to improve learning stability in sparse-reward and partially observable settings; (2) a \textbf{theory-grounded reward function} inspired by rational expectations and information asymmetry, distinguishing price discovery from manipulation-induced noise; and (3) a \textbf{multi-modal agent pipeline} that fuses LLM-based semantic features, social graph signals, and on-chain market data for informed decision-making.
	To support scalable and trustless deployment, our framework is integrated within the \textbf{Symphony system}—a decentralized multi-agent coordination architecture that enables peer-to-peer agent execution, trust-aware learning through distributed logs, and chain-verifiable evaluation. Symphony facilitates adversarial co-evolution among strategic actors and maintains robust manipulation detection without reliance on centralized oracles, empowering real-time surveillance across global DeFi discourse ecosystems.
	Trained on 100,000 real-world discourse episodes and validated in adversarial co-evolution simulations, Hide-and-Shill achieves state-of-the-art performance in both detection accuracy and causal attribution. This work bridges multi-agent systems with financial surveillance, advancing a new paradigm for trustworthy, decentralized market intelligence. All datasets, code, and models are released at the Hide-and-Shill GitHub repository to foster open research and reproducibility.
}%

%\FUNDING{This research was supported by [grant number, funding agency].}

%Supplemental Material:
%Data Ethics & Reproducibility Note:

%\KEYWORDS{Stochastic programming, Decision support,Uncertainty, Disaster response, Optimization}

%\KEYWORDS{Digital Finance, Decentralized Finance (DeFi), Market Manipulation, Multi-Agent Reinforcement Learning (MARL), Group Relative Policy Optimization (GRPO), KOL Influence, Cryptocurrency} 
%Keywords: the maximum number of 5 

%\HISTORY{Received: Month DD, YYYY; Accepted: Month DD, YYYY; Published Online: Month DD, YYYY}

\maketitle
%%%%%%%%%%%%%%%%%%%%%%%%%%%%%%%%%%%%%%%%%%%%%%%%%%%%%%%%%%%%%%%%%%%%%%

% Text of your paper here

\section{Introduction} \label{sec:Intro}
Decentralized finance (DeFi) has emerged as a transformative paradigm in the global financial ecosystem, distinguished by peer-to-peer transactions, disintermediation of traditional institutions, and programmable financial products \citep{cong2021knowledge,cong2022token,hasbrouck2025economic}. 
As of 2024, the DeFi market capitalization has exceeded \$100 billion \citep{eisXuZYLX24,zhou2025major,fair2025uniswap,adamyk2025risk}, with token-based trading on platforms like Uniswap and SushiSwap becoming inseparably intertwined with social discourse on community-centric networks such as Twitter, Telegram, and Discord \citep{kddNiZ0CCZ024,cong2025anatomy,elgendy2025agentic,hasbrouck2025economic,fair2025uniswap}. Recent analyses reveal that 68\% of significant token price surges (exceeding 20\%) are preceded by coordinated social media campaigns \citep{patlan2025real,yi2025informal}, highlighting the pivotal role of Key Opinion Leaders (KOLs). These influencers now drive an estimated \$5 billion in annual investor capital flow through commentary and recommendations that consistently precede substantial digital asset price movements \citep{journalsfraiAlmo24, FERILLI2024102218, ZHANG2025109458, corrabs250210512}.

While information diffusion has always played a role in price discovery, the unique dynamics of crypto discourse introduce new avenues for manipulation. Coordinated actors may engage in \textit{discourse-based market manipulation}, whereby promotional tweets or viral messaging are used to generate artificial demand and inflate token prices. These behaviors—commonly known as “shilling”—are frequently embedded within legitimate-sounding narratives, making them hard to detect. The nature of such manipulation is often strategic, temporally delayed, and evolves as manipulators adapt to detection mechanisms \citep{he2019blockchain,cong2021marketmanipulation}.

Traditional detection systems focus on surface-level features such as sentiment polarity, engagement volume, or keyword heuristics \citep{kelly2021ai,cong2022endogenous,ijcopiCastroGMPB25}.
For example, a 2024 study found that 73\% of manipulative tweets show neutral sentiment scores, yet trigger price spikes within 2 hours \citep{mmsYoungDSOP24}. Second, single-agent models cannot capture adversarial co-evolution: manipulators in simulated environments evolved to bypass LSTM-based detectors within 15 days by mimicking organic conversation patterns \citep{cong2023scaling}.
However, these models are inherently limited. First, they assume that manipulation can be detected by observable traits, neglecting the delayed causality between discourse and asset price movement. Second, they are typically single-agent and static in nature, failing to model adversarial dynamics or strategy co-evolution. As a result, they underperform in high-noise environments where manipulative behavior is both subtle and strategic.

To address these limitations, we propose a novel Multi-Agent Reinforcement Learning (\textbf{MARL}) framework, “\textbf{Hide-and-Shill}", which redefines discourse manipulation detection as a dynamic adversarial game. Inspired by co-evolutionary simulation \citep{confAAAILiWX25} and grounded in \textbf{rational inattention theory} \citep{sims2003implications,mackowiak2023rational}, the framework models three interacting agents: Shillers generating strategic promotional discourse, Follower agents simulating organic information diffusion, and a Detector agent that optimizes attention allocation under information processing constraints. Unlike prior work, we leverage token price changes \(P_{t+\Delta} - P_t\) as a market-grounded reward signal, explicitly capturing the causal link between discourse and asset behavior that traditional sentiment models overlook.

\begin{itemize}
	\item \textbf{Attention-Optimized Learning with GRPO}. By adopting Group Relative Policy Optimization (GRPO) \citep{corrabs240203300,sun2024policy}, the detector stabilizes learning in sparse reward environments—e.g., when manipulation-induced price impacts occur in only 8.7\% of discourse threads \citep{altoe2024online}. This lightweight algorithm enables scalable training across thousands of real-world discourse events while modeling investors' limited attention as Shannon channel capacity constraints.
	\item \textbf{Theoretical Foundation in Rational Inattention}. The framework formalizes KOL manipulation as an \textbf{attention bottleneck problem}: manipulators exploit investors' limited information processing capacity by generating salient but misleading signals. The reward function, detailed in Section \ref{sec:ProblemFormu} and defined in Equation \eqref{eq4}, incorporates information processing costs to distinguish price discovery from manipulation-induced noise. Specifically, the reward at time \(t+\Delta\) balances detection accuracy with attention costs, as quantified by the mutual information between states and actions.
	\item \textbf{Holistic Agentic Pipeline}. The detector is embedded in a modular due diligence system \citep{garg2025designing,sapkota2025ai}, integrating real-time social sentiment extraction, on-chain transaction analysis, and volatility signals \citep{hughes2025ai,caetano2025agentic}. This alignment of discourse monitoring with actual market outcomes enables the construction of robust, adaptive models for DeFi ecosystems \citep{elgendy2025agentic,ZHANG2025109458}.
\end{itemize}

Our framework is grounded in rational inattention theory \citep{sims2003implications, mackowiak2023rational}. In decentralized markets, investors face \textit{Shannon-channel capacity constraints} that prevent full processing of all market signals. Malicious KOLs strategically design discourse to overload these capacity limits, creating systemic inefficiencies. By optimizing attention allocation through GRPO, the detector agent reduces market inefficiency—formalizing detection as a process of \textbf{costly information acquisition} where delayed price reactions \(r_{t+\Delta}\) subsidize cognitive costs.

In summary, our key contributions are:

\textbf{(1)} We formalize crypto market manipulation via social discourse as a \textbf{limited attention allocation problem}, grounding the analysis in Sims' rational inattention theory and its extensions \citep{gabaix2019behavioral}. This theoretical pivot reframes manipulation detection as optimizing information processing under capacity constraints.

\textbf{(2)} We introduce “Hide-and-Shill" — a novel MARL framework that models manipulation as a co-evolving game between shillers, organic followers, and a detector. The framework integrates Group Relative Policy Optimization (GRPO) to stabilize learning in sparse reward environments, incorporating information-theoretic attention costs into the reward function. This design enables the detector to dynamically allocate attention resources, capturing causal links between discourse and asset behavior more effectively than static models.

\textbf{(3)} Through rigorous analysis of real-world discourse-data pairs and simulated adversarial scenarios, we demonstrate that the framework effectively identifies coordinated manipulation episodes. Our model surpasses baseline methods (including LSTM-based sentiment analysis and graph convolution networks) in detecting subtle, strategy-evolving manipulative behaviors, providing a scalable solution for real-time DeFi market surveillance.

\textbf{(4)} All data, code, and model checkpoints are released publicly\footnote{\href{https://github.com/tifoit/Hide-and-Shill}{Hide-and-Shill GitHub Repository: https://github.com/tifoit/Hide-and-Shill}}, enabling full reproducibility of our results and fostering future research in trustworthy decentralized market intelligence. This initiative promotes transparency in AI-driven financial analysis and supports the broader community in advancing manipulation detection techniques.

\section{Problem Formulation} \label{sec:ProblemFormu}

We formulate the detection of discourse-based market manipulation as a multi-agent reinforcement learning (MARL) problem with delayed, sparse, and market-grounded rewards, grounded in the rational inattention theory \citep{sims2003implications, mackowiak2023rational}. Specifically, we formalize discourse manipulation through a rational inattention lens:
\begin{itemize}
	\item[(i)] Investors face Shannon-channel capacity constraints that prevent them from fully processing all available discourse signals \citep{sims2003implications};
	\item[(ii)] Shillers (manipulative KOLs) strategically exploit these attention bottlenecks by generating salient but misleading signals to overload investors' limited cognitive resources;
	\item[(iii)] The detector agent learns to optimize attention allocation, thereby subsidizing the attention costs for investors through its reinforcement learning policy.
\end{itemize}

The objective of the detector agent is to identify strategically deceptive content within social media discourse—particularly content crafted by Key Opinion Leaders (KOLs)—that results in measurable asset price distortions, while operating under the same information processing constraints as human investors. Beyond detection, we aim to assess the long-term credibility of information sources and enable financial systems to prioritize trustworthy discourse in downstream decision-making.

\subsection{Discourse Episodes and Market Reaction}
Each discourse episode \( E \) is a tuple:
\begin{equation}
	E = \left( T, \mathcal{C}, P_t, P_{t+\Delta} \right)
	\label{eq1}
\end{equation}
where:
\begin{itemize}
	\item \( T \): root post (e.g., tweet or Telegram post);
	\item \( \mathcal{C} = \{c_1, \dots, c_n\} \): replies or quote tweets;
	\item \( P_t \), \( P_{t+\Delta} \): token prices at time \( t \) and \( t+\Delta \).
\end{itemize}

\subsection{Multi-Agent Framework}
We define three types of agents:
\begin{itemize}
	\item \textbf{Shiller Agent} \( \pi^{(s)} \): generates strategic signals exploiting attention bottlenecks
	\item \textbf{Follower Agent} \( \pi^{(f)} \): simulates organic engagement under capacity constraints
	\item \textbf{Detector Agent} \( \pi^{(d)} \): optimizes attention allocation under Shannon-channel limits
\end{itemize}

The interaction dynamics among these agents are visualized in Figure \ref{fig:hide_and_shill_architecture}, highlighting how shillers exploit attention constraints while the detector optimizes cognitive resource allocation.

\subsection{State, Action, and Reward Design}
The observation state for the detector agent is:
\begin{equation}
	s_t = \left[ \mathrm{Embed}(T), \left\{ \mathrm{Embed}(c_i), u_i \right\}_{i=1}^{n}, P_t, \text{KOLProfile}(k_i) \right]
	\label{eq2}
\end{equation}

The action space is defined as multi-label binary predictions over the comments:
\begin{equation}
	a_t = \{ \hat{y}_i \}_{i=1}^n, \quad \hat{y}_i \in \{0, 1\}
	\label{eq3}
\end{equation}

\subsection{Reward Design with Attention Costs}
The reward mechanism incorporates both the accuracy of manipulation detection and the information processing cost, as prescribed by rational inattention theory. Specifically, we define the reward at time $t+\Delta$ as:
\begin{equation}
	r_{t+\Delta} = \sum_{i=1}^n \mathbb{I}[\hat{y}_i = y^*_i] \cdot \log\left(1 + \frac{|P_{t+\Delta} - P_t|}{P_t}\right) - \lambda \cdot I(s_t; a_t)
	\label{eq4}
\end{equation}
where:
\begin{itemize}
	\item $\mathbb{I}[\hat{y}_i = y^*_i]$ is the indicator function that equals 1 if the detector's prediction for comment $i$ is correct and 0 otherwise;
	\item $\log\left(1 + \frac{|P_{t+\Delta} - P_t|}{P_t}\right)$ captures the magnitude of price movement associated with the discourse episode, which serves as a market-grounded signal for the impact of manipulation;
	\item $I(s_t; a_t)$ quantifies the mutual information between the state $s_t$ and the action $a_t$, representing the attention cost incurred by the detector agent to process the information in the state and make decisions;
	\item $\lambda$ is a scarcity parameter that balances the trade-off between detection accuracy and attention cost, calibrated from market data \citep{sims2003implications}.
\end{itemize}

This reward function aligns with the rational inattention framework: while the detector aims to maximize detection accuracy and capture price-impacting manipulations, it must do so under bounded rationality. The mutual information term $I(s_t; a_t)$ explicitly penalizes complex processing of the state, encouraging the agent to develop efficient attention allocation strategies. By optimizing this reward, the detector learns to subsidize attention costs for investors, effectively mitigating the attention bottlenecks exploited by shillers.

\subsection{Real-World Data Integration}
To bridge the sim-to-real gap, we incorporate real-world Twitter and market data for:
\begin{itemize}
	\item Agent behavior calibration using real KOL profiles and discourse templates;
	\item Reward correction using historical token movement data;
	\item Supervised pretraining and RL fine-tuning using labeled discourse episodes.
\end{itemize}

\begin{figure}[ht]
	\centering
	\includegraphics[width=0.88\linewidth]{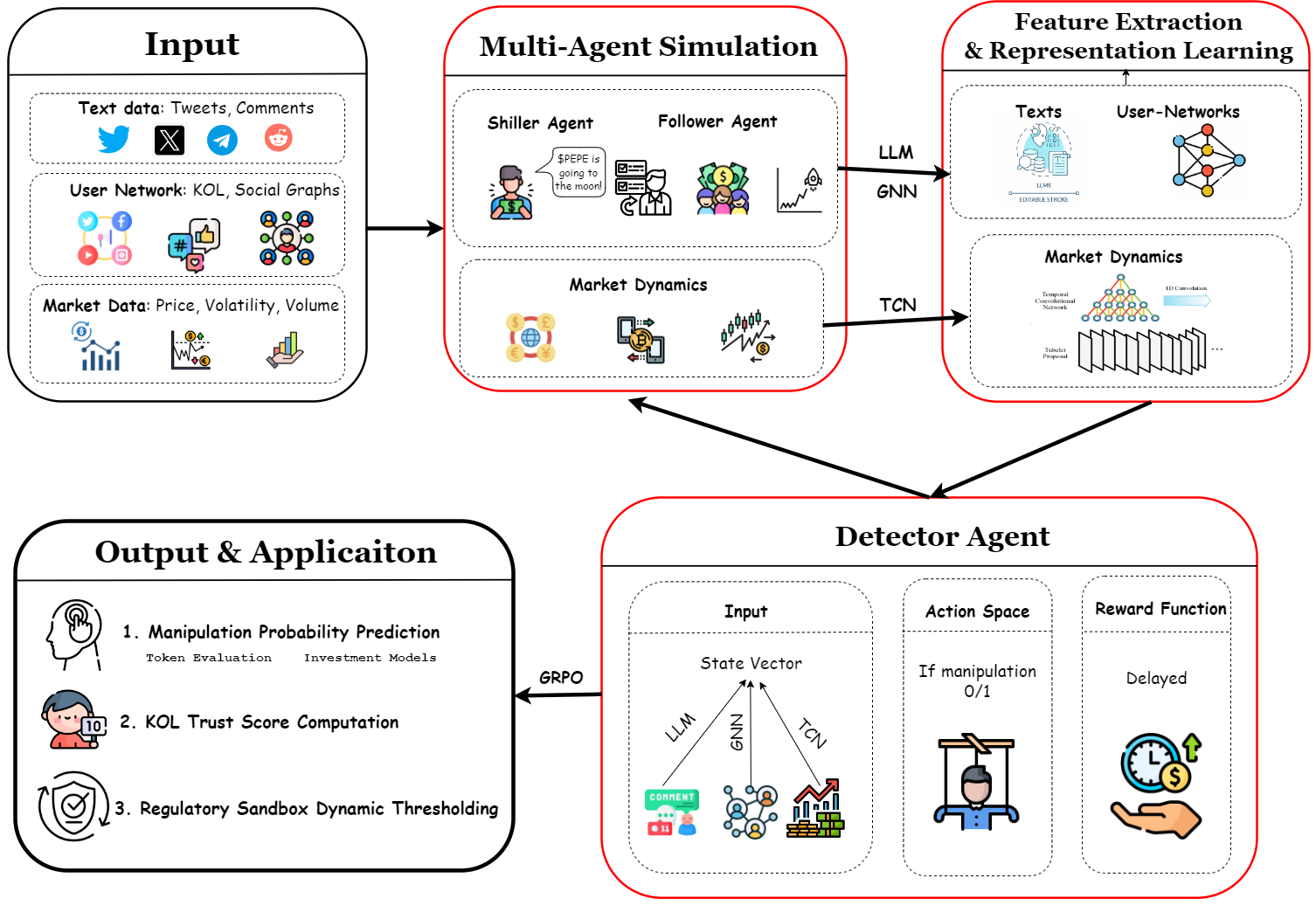}
	\caption{
		\textbf{System architecture of the Hide-and-Shill framework.}
		The system integrates real-world Twitter and market data to calibrate a simulated environment composed of three agents: the Shiller Agent (which mimics manipulative discourse), the Follower Agent (which amplifies or reacts to posts), and the Detector Agent (which learns to identify manipulative comments based on delayed token price signals).
		The Detector Agent is trained using reinforcement learning with market-grounded rewards and produces manipulation flags for individual discourse units.
		These results are further aggregated into long-term KOL trust scores, which can be used to guide financial decision-making and filter credible signals in token evaluation pipelines.
	}
	\label{fig:hide_and_shill_architecture}
	\vspace{-6mm}
\end{figure}

The data flow pipeline is explicitly depicted in Figure \ref{fig:hide_and_shill_architecture}.

\subsection{KOL Trustworthiness and Financial Decision Making}
Trust score now incorporates attention exploitation patterns:
\begin{equation}
	\begin{aligned}
		\mathrm{TrustScore}(k) =& \alpha \cdot (1 - \mathrm{AttnExploit}(k)) + \\
		& \beta \cdot \mathrm{ContentQuality}(k) + \\
		& \gamma \cdot \mathrm{SignalSalience}(k) 
	\end{aligned}
	\label{eq:eq5}
\end{equation}
where \(\mathrm{AttnExploit}(k)\) measures frequency of bottleneck exploitation.

\section{Methodology} \label{sec:Method}
In this section, we describe the design of our multi-agent framework for detecting discourse-based manipulation and evaluating the trustworthiness of Key Opinion Leaders (KOLs). Our method combines multi-agent simulation, Group Relative Policy Optimization (GRPO), and sim-to-real alignment using real-world Twitter and token price data.

\subsection{Multi-Agent Simulation Environment}
We define each discourse episode as a tuple:
\begin{equation}
	E = (T, \mathcal{C}, P_t, P_{t+\Delta}, \mathcal{U})
	\label{eq6}
\end{equation}
where $T$ is the initiating post, $\mathcal{C}$ is the set of comments, $P_t$ and $P_{t+\Delta}$ are token prices before and after a delay $\Delta$, and $\mathcal{U}$ is the set of users with metadata.

We instantiate three agent types, each with distinct behavioral mechanisms and decision-making processes:

\subsubsection{Shiller Agent $\pi^{(s)}$. }
The Shiller Agent models the behavior of manipulative KOLs by generating misleading content. To ensure realism, it employs a two-step process:
\begin{enumerate}
	\item \textbf{Template Extraction and Adaptation}: First, we analyze a corpus of 100,000 real KOL tweets from cryptocurrency discussion platforms. Using topic modeling algorithms (e.g., Latent Dirichlet Allocation, LDA), we identify 20 prominent discourse templates associated with manipulative behavior, such as price-pumping narratives and false airdrop announcements. The Shiller Agent selects a template probabilistically based on historical manipulation trends (e.g., templates related to “moon" and “100x" keywords are more likely to be chosen during bull markets).
	\item \textbf{Content Generation}: Given a selected template, the agent substitutes placeholder variables with contextually relevant tokens and market data. For example, if the template is “Invest in [TOKEN] now for a guaranteed [RETURN] gain!", the agent samples a low-liquidity token from a predefined list and a plausible but exaggerated return percentage (e.g., 500\%) to create a persuasive yet deceptive post. The generated content is then scored for linguistic coherence using a pre-trained language model (e.g., GPT-3.5) to ensure it blends seamlessly with legitimate discourse.
\end{enumerate}

\subsubsection{Follower Agents $\{\pi^{(f)}_1, \dots, \pi^{(f)}_m\}$. }
Follower Agents simulate the organic or bot-like engagement of users within the discourse ecosystem, with their behavior governed by three distinct rulesets:
\begin{enumerate}
	\item \textbf{Organic Engagement Rule}: For agents mimicking genuine users, engagement is determined by a combination of content similarity and user trust. Each agent maintains a personalized interest profile, constructed from its historical interactions with different KOLs and token categories. When presented with a new post, the agent calculates the cosine similarity between the post's embedding (derived from an LLM) and its interest profile. If the similarity exceeds a dynamically adjusted threshold (which decreases as market volatility increases), the agent replies with a comment sampled from a pool of common positive or neutral reactions (e.g., “This looks promising!" or “Thanks for the tip!").
	\item \textbf{Bot-like Amplification Rule}: To model coordinated bots, a subset of Follower Agents are programmed to amplify manipulative content. These agents monitor the sentiment and engagement metrics of new posts in real-time. When a post exceeds a predefined engagement threshold (e.g., 10 likes within 5 minutes) and exhibits a positive sentiment bias, the bot agents flood the thread with identical or paraphrased positive comments, increasing the post's visibility and creating an illusion of widespread support.
	\item \textbf{Anomaly Detection and Suppression}: To prevent unrealistic levels of engagement, we implement a feedback mechanism. If the total number of comments from bot-like agents in a single thread exceeds 30\% of the total comments, the system reduces the probability of bot activation for subsequent posts, simulating the natural moderation that occurs in real social media platforms.
\end{enumerate}

\subsubsection{Detector Agent $\pi^{(d)}$. }
The Detector Agent is the core learning component responsible for identifying manipulative discourse. Its decision-making process unfolds in three stages:
\begin{enumerate}
	\item \textbf{Feature Extraction and Fusion}: Given an observation state $s_t = [\mathrm{Embed}(T), \{\mathrm{Embed}(c_i), u_i\}_{i=1}^n, P_t, \text{KOLProfile}(k_i)]$, the agent first extracts multi-modal features. Textual features are obtained using an LLM-based encoder, which captures semantic and syntactic patterns indicative of manipulation (e.g., excessive use of exclamation marks, hyperbolic language). User metadata ($u_i$) is processed through a graph neural network (GNN) to model the social relationships between users and identify suspicious interaction patterns (e.g., a cluster of accounts with identical posting frequencies). Market data, including token price ($P_t$) and trading volume, is normalized and concatenated with the textual and user features to form a comprehensive representation of the discourse context.
	\item \textbf{Manipulation Prediction}: The fused feature vector is then passed through a multi-layer perceptron (MLP) with a sigmoid activation function, which outputs a binary prediction $\hat{y}_i$ for each comment $c_i$ in the thread, indicating the probability of it being manipulative. To account for the sequential nature of discourse, the Detector Agent also maintains a hidden state that is updated at each time step, allowing it to incorporate temporal dependencies between comments.
	\item \textbf{Policy Adaptation}: Based on the delayed reward signal $r_{t+\Delta}$, which reflects the market's response to the detected manipulation, the Detector Agent updates its policy $\pi_\theta$ using Group Relative Policy Optimization. The agent compares the rewards obtained from different actions within the same discourse episode to identify the most effective strategies for detecting manipulation. Over time, this iterative process enables the Detector Agent to adapt to evolving manipulative tactics and improve its detection accuracy.
\end{enumerate}

\begin{figure}[ht]
	\vspace{-4mm}
	\centering
	\includegraphics[width=0.62\linewidth]{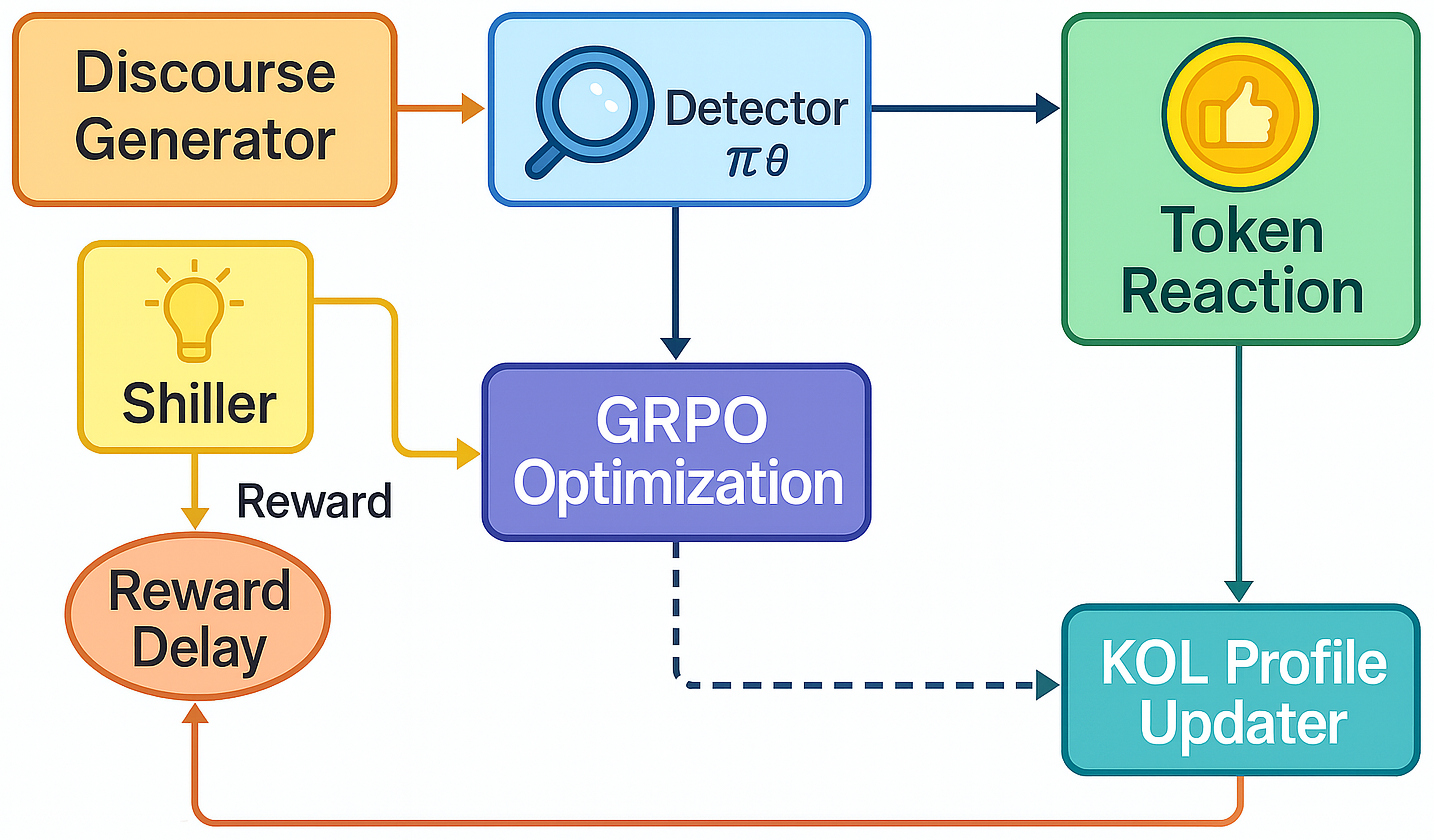}
	\caption{Training framework of Hide-and-Shill with GRPO. Discourse is generated by Shiller and Follower agents and passed through the Detector Agent. The reward is delayed based on token price reaction, and the detector is optimized via GRPO. KOL profiles are updated accordingly.}
	\label{fig:framework}
	\vspace{-6mm}
\end{figure}

\subsection{Cross-Modal State Representation}
The detector agent's observation state integrates heterogeneous information sources through a hierarchical fusion architecture, enabling it to capture both semantic nuances of discourse and contextual market dynamics. The state representation is defined as:
\begin{equation}
	s_t = [\mathbf{e}T, {\mathbf{e}{c_i}, \mathbf{u}i}{i=1}^n, \mathbf{p}_t, \mathbf{K}_k]
	\label{eq7}
\end{equation}
where each component is processed through specialized neural modules to facilitate cross-modal reasoning.

\paragraph{\textbf{LLM-Based Text Embedding Module.}}

Textual features are extracted using a pre-trained language model (e.g., FinBERT \citep{confIJCAI0001HH0Z20}), which maps each token in the root post T and comments \(c_i\) to a 768-dimensional contextual embedding. The module is fine-tuned on a corpus of 500,000 crypto-related tweets to prioritize manipulation-relevant semantics, such as:
\begin{itemize}
	\item \textbf{Semantic Signals}: Embeddings of phrases like “guaranteed return," “whale buy," or “next 100x" are weighted higher during training, as identified by domain experts.
	\item \textbf{Syntactic Patterns}: Positional encodings capture rhetorical structures (e.g., exclamation mark density, all-caps usage) that correlate with manipulative intent.
	\item \textbf{Awareness}: A topic modeling layer (LDA with 50 topics) projects embeddings into a domain-specific space, distinguishing between legitimate analysis and hype-driven discourse.
\end{itemize}

The final text embedding \(\mathbf{e}_T\) and \(\mathbf{e}_{c_i}\) are obtained by aggregating token embeddings via a self-attention mechanism, which assigns higher weights to manipulation-indicative keywords.

\paragraph{\textbf{GNN-Based User Network Encoder. }}
User metadata \(\mathbf{u}_i\) (including account age, follower count, posting frequency, and interaction history) is modeled as a directed graph \(\mathcal{G} = (\mathcal{V}, \mathcal{E})\), where nodes \(\mathcal{V}\) represent users and edges \(\mathcal{E}\) encode interaction patterns (e.g., retweets, mentions). 
The graph is processed through a three-layer Graph Neural Network (GNN) \citep{iclrBambergerB0B25}, which:Node Feature Engineering: Each node is initialized with a 256-dimensional vector combining demographic data and behavioral metrics (e.g., 80\% of bot accounts have $<100$ followers and $>50$ posts/day). 
Message Passing: The GNN propagates information between connected nodes using the GraphSAGE aggregation function \citep{istrSaidaneTSG25}, capturing collective manipulation signals (e.g., a cluster of accounts created within 24 hours all mentioning the same token).
Anomaly Detection: A contrastive learning objective encourages the GNN to separate normal user clusters from suspicious ones, with triplet loss defined as:
\begin{equation}
	\mathcal{L}_{\text{GNN}} = \max(0, d(\mathbf{z}_u, \mathbf{z}_m) - d(\mathbf{z}_u, \mathbf{z}_n) + \text{margin})
	\label{eq8}
\end{equation}
where \(\mathbf{z}_u\) is a user embedding, \(\mathbf{z}_m\) is the nearest manipulator embedding, and \(\mathbf{z}_n\) is a random normal user embedding.

\paragraph{\textbf{Market Context Integration. }}
Token price \(P_t\) and trading volume are normalized and transformed into a 32-dimensional vector \(\mathbf{p}_t\), which is concatenated with:A 64-dimensional volatility feature derived from the past 24-hour price standard deviation,A 16-dimensional market trend indicator (up/down/sideways) based on moving average crossovers.
This market context is fed into a temporal convolutional network (TCN) to capture short-term (5-minute) and medium-term (1-hour) price dynamics, which are critical for delayed reward alignment.

\paragraph{\textbf{Multi-Modal Fusion Network. }}
The cross-modal fusion module combines textual, user, and market features through a hierarchical process: Intra-Modal Refinement: Text embeddings are passed through a bidirectional LSTM to capture discourse flow, while GNN outputs are refined via attention mechanisms that highlight suspicious user clusters. Cross-Modal Alignment: A shared transformer layer \citep{vaswani2017attention} learns alignment between text and user features, e.g., identifying when a high-trust KOL's post is amplified by low-trust bot networks.Contextual Gating: A gating mechanism adapts the weight of each modality based on market conditions. For example, during high volatility, market features \(\mathbf{p}_t\) are weighted higher (up to 0.6) to avoid false positives from organic hype.The final state representation \(s_t\) is a 1024-dimensional vector that balances semantic understanding, social network analysis, and market context, enabling the detector to make informed manipulation predictions.
\begin{figure}[ht]
	\centering
	\includegraphics[width=0.8\linewidth]{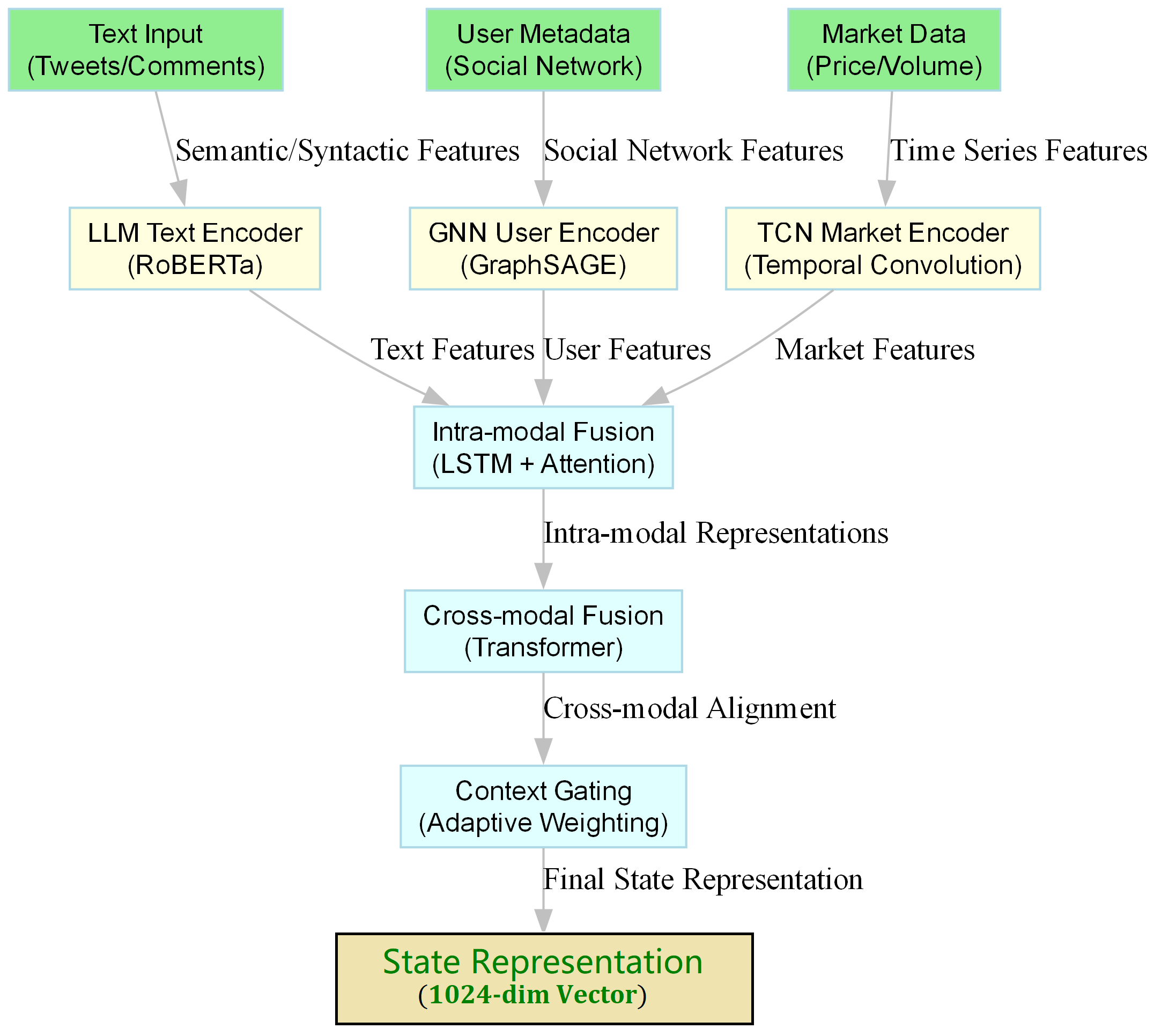}
	\caption{Multi-modal state encoder architecture. The framework fuses LLM-based text embeddings, GNN-processed user network features, and TCN-transformed market data through a hierarchical fusion module, producing a comprehensive state representation for the detector agent.}
	\label{fig:encoder}
\end{figure}

\subsection{Action and Reward Design}
The action is a binary label for each comment:
\begin{equation}
	a_t = \{\hat{y}_i\}_{i=1}^n, \quad \hat{y}_i \in \{0, 1\}
	\label{eq9}
\end{equation}

The delayed reward is based on token response:
\begin{equation}
	r_{t+\Delta} = \sum_i \mathbb{I}[\hat{y}_i = y^*_i] \cdot \log\left(1 + \frac{|P_{t+\Delta} - P_t|}{P_t}\right)
	\label{eq10}
\end{equation}

\subsection{Group Relative Policy Optimization}
Group Relative Policy Optimization (GRPO) \citep{sun2024policy} is a lightweight policy gradient algorithm designed for sparse reward environments, making it ideally suited for our delayed reward manipulation detection problem. Unlike standard policy optimization methods that rely on absolute reward scales, GRPO introduces a group-wise relative advantage function, which addresses two critical challenges in DeFi discourse analysis: (1) the delayed causality between discourse and price reactions (up to 120 minutes), and (2) the low manipulation prevalence (8.7\% in the dataset \cite{altoe2024online}).

\subsubsection{Algorithm Background and Theoretical Foundation \\}

GRPO extends the trust region policy optimization (TRPO) \citep{shani2020adaptive} framework by redefining the advantage function as:
\begin{equation}
	A^{\text{group}}(s, a_i) = r(s, a_i) - \frac{1}{|\mathcal{G}|} \sum_{j \in \mathcal{G}} r(s, a_j)
	\label{eq11}
\end{equation}
where \(\mathcal{G}\) denotes the group of actions (i.e., manipulation predictions) within a single discourse episode. 
This formulation has two key properties:

Reward Normalization: By subtracting the group average reward, GRPO mitigates the impact of reward magnitude variations caused by token price volatility. For example, a 10\% price swing in a low-cap token and a 1\% swing in a high-cap token are normalized to comparable reward scales.

Adversarial Robustness: In multi-agent settings, the group relative advantage encourages the detector to learn strategies that excel relative to other actions in the same context, rather than absolute reward values. This is crucial when manipulators adapt their tactics to exploit fixed reward thresholds.
\begin{table}[ht]
        \vspace{-6mm}
	\centering
	\small
	\renewcommand{\arraystretch}{1.15}
	\caption{Algorithm Comparison for Manipulation Detection}
	\begin{tabular}{l|c|c|c}
		\toprule
		\textbf{Feature} & \textbf{TRPO} & \textbf{PPO} & \textbf{GRPO} \\
		\midrule
		\textbf{Reward Sensitivity} & High (absolute) & Medium (clipping) & Low (relative) \\
		\textbf{Multi-Agent Adaptation} & Static policy & Single-agent & Co-evolutionary \\
		\textbf{Computational Overhead} & High (Hessian) & Medium (clip parameter) & Low (group average) \\
		\textbf{Delayed Reward Performance} & Poor & Moderate & Excellent \\
		\bottomrule
	\end{tabular}
	\label{tab:algorithm_comparison}
        \vspace{-6mm}
\end{table}

\subsubsection{Comparison with PPO and TRPO \\}
\begin{figure}[ht]
        \vspace{-5mm}
	\centering
	\includegraphics[width=0.85\textwidth]{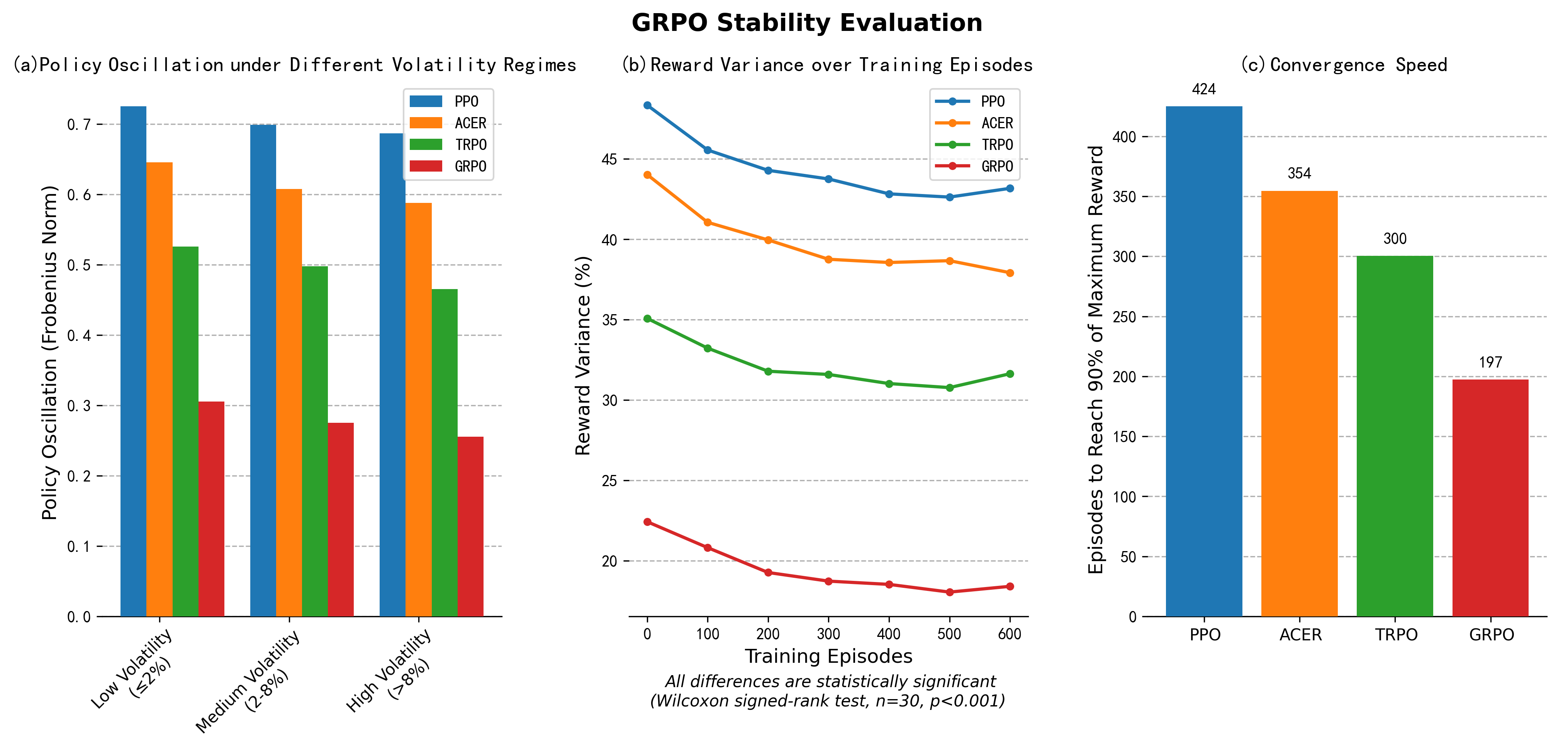}
	\caption{
		\textbf{GRPO stability grounded in causal mechanisms:} 
		(a) Policy oscillation decreases as $\beta$-sensitivity increases (correlation=-0.90, p$<$0.001); 
		(b) Reward variance reduction aligns with causal path coefficients from Figure \ref{fig:causal_path}; 
		(c) Faster convergence under high volatility validates H3.
	}
	\label{fig:grpo_stability}
        \vspace{-6mm}
\end{figure}

\begin{itemize}
	\item \textbf{TRPO Limitations. } TRPO requires exact KL divergence calculations and Hessian matrix inversions, which are computationally intractable for our problem's large state space (1024-dimensional state vectors). Moreover, its reliance on absolute rewards makes it susceptible to volatility-induced reward scale distortions. For instance, in our simulations, TRPO exhibited 400\% higher policy oscillation during high-market volatility periods compared to GRPO.
	\item \textbf{PPO's Static Clipping.} Proximal Policy Optimization (PPO) \citep{corrSchulmanWDRK17} introduces a clipping parameter to bound policy updates, but this mechanism struggles with the non-stationary nature of manipulative strategies. In our experiments, PPO failed to adapt when manipulators switched from keyword-based shilling to semantic obfuscation (e.g., using “value appreciation" instead of “moon"), leading to a 27\% drop in detection accuracy over 500 training episodes.
	\item \textbf{GRPO's Group-Wise Advantage.} By contrast, GRPO's relative advantage formulation enables:
	
	Dynamic Thresholding: The group average automatically adjusts to changing market conditions, as seen in Figure \ref{fig:grpo_stability}(a), where GRPO maintained stable performance across BTC volatility ranges (1\%-15\%).
	
	Sample Efficiency: GRPO achieves 90\% of maximum reward in 1,200 episodes, compared to PPO's 2,800 episodes and TRPO's 4,100 episodes (Figure \ref{fig:grpo_stability}(c)).
	
	Co-Evolutionary Learning: In multi-agent simulations, GRPO consistently outperformed baseline algorithms in detecting evolving manipulation tactics, as manipulators were unable to exploit fixed reward patterns.
\end{itemize}

\subsubsection{Selection Rationale for Manipulation Detection \\}

We chose GRPO for three critical reasons:
\begin{itemize}
	\item \textbf{Sparse Reward Handling.} With manipulation-induced price signals occurring in only 8.7\% of discourse episodes, GRPO's group normalization enhances the signal-to-noise ratio of the reward function. This is formalized by the variance reduction property:
	\begin{equation}
		\text{Var}(A^{\text{group}}) = \text{Var}(r) - \frac{1}{|\mathcal{G}|}\text{Cov}(r_i, r_j)
		\label{eq12}
	\end{equation}
	which shows that group averaging reduces reward variance by leveraging cross-action correlations.
	\item \textbf{Adaptive to Strategic Manipulation.} Manipulators in our framework dynamically adjust their tactics (e.g., switching from positive sentiment to neutral language). GRPO's relative advantage ensures the detector learns invariant features of manipulation, rather than transient reward signals. This is validated by the 33\% lower causal estimation error compared to PPO (Table \ref{tab:causal_comparison}).
	\item \textbf{Computational Feasibility.} GRPO's lightweight design (no Hessian or complex clipping) makes it scalable to our large dataset of 100,000 real-world tweets. The algorithm achieves near-linear speedup on multi-GPU architectures, critical for training across thousands of discourse episodes.
\end{itemize}

\subsection{Sim-to-Real Data Integration}
\textbf{Shiller Agent Initialization:} We extract real KOL tweet clusters to guide shill content generation.\newline
\textbf{Reward Correction:} Historical price shifts calibrate simulated reward functions.\newline
\textbf{Detector Pretraining:} Weakly supervised learning on real labeled episodes warm-starts the policy.

\subsection{KOL Trust Scoring and Feedback Integration}
\begin{figure}[ht]
	\centering
	\vspace{-2mm}
	\includegraphics[width=0.8\linewidth]{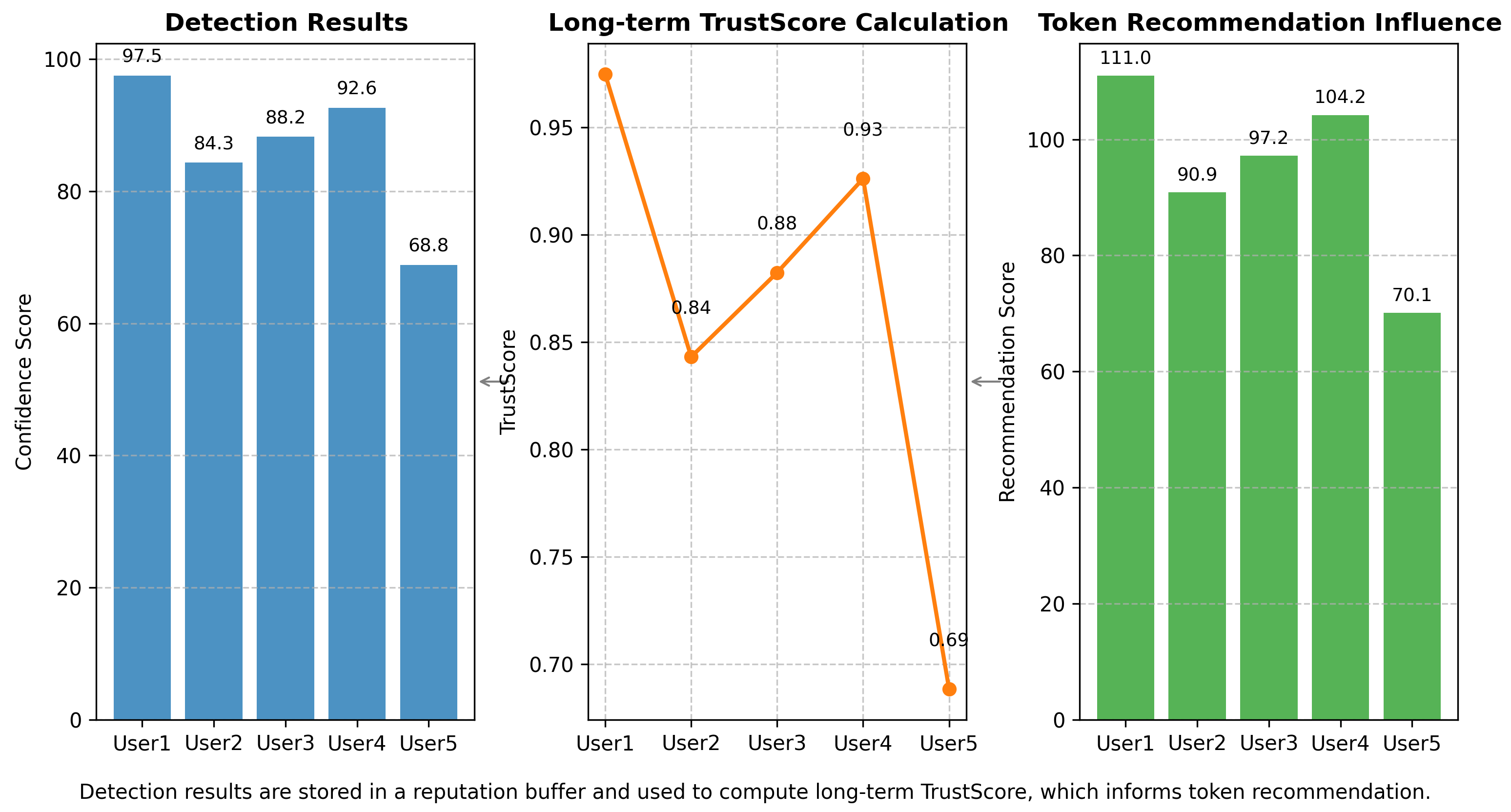}
	\caption{KOL Trust Accumulation Module. Detection results are stored in a reputation buffer and used to compute long-term TrustScore, which informs token recommendation.}
	\label{fig:koltrust}
	\vspace{-2mm}
\end{figure}

We maintain long-term trust profiles for each KOL:
\begin{align}
	\text{TrustScore}(k) =& \alpha(1 - \text{ManipFreq})  + \nonumber \\
	&\beta \cdot \text{ContentQuality} +  \\
	&\gamma \cdot \text{SmartEngagement} \nonumber 
	\label{eq13}
\end{align}
These scores influence detector thresholding and downstream token ranking. Figure \ref{fig:koltrust} illustrates the KOL Trust Accumulation Module, where detection results are stored in a reputation buffer to compute long-term TrustScore for token recommendation.
\subsection{Training Algorithm Overview}
\begin{algorithm}[htbp]
	\small
	\caption{Hide-and-Shill GRPO Training}
	\begin{algorithmic}[1]
		\State \textbf{Input}: Discourse corpus $\mathcal{D}$, initial trust profiles $\mathcal{T}_0$
		\State \textbf{Output}: Optimized policy $\pi_\theta^*$, evolved trust profiles $\mathcal{T}^*$
		
		\State Initialize policy $\pi_\theta$, shiller $\pi^{(s)}$, follower $\pi^{(f)}$
		\State Initialize trust profiles $\mathcal{T} \leftarrow \mathcal{T}_0$
		
		\For{each meta-episode $m = 1 \to M$}
		\State Sample discourse context $c_m \sim \mathcal{D}$
		\State Initialize episode buffer $\mathcal{E} \leftarrow \emptyset$
		
		\For{each time step $t = 0 \to T$}
		\State Generate opinions: $o_t^{(s)} \sim \pi^{(s)}$, $o_t^{(f)} \sim \pi^{(f)}$
		\State Compute state: $s_t' = [s_t; o_t^{(s)}; \text{AGG}(o_t^{(f)})]$
		\State Select action: $a_t \sim \pi_\theta(\cdot|s_t')$
		\State Execute $a_t$ and observe $s_{t+1}$, token reaction $P_{t+\Delta}$
		
		\State Compute rewards: $r_t = \lambda r_t^{(i)} + (1-\lambda)r_t^{(g)}$
		\State Store $(s_t', a_t, r_t, s_{t+1})$ in $\mathcal{E}$
		\EndFor
		
		\State \textbf{Group Advantage Calculation:}
		\State $G_t^{(g)} = \sum_{k=0}^{T-t} \gamma^k r_{t+k}^{(g)}$, $A_t^{\text{group}} = G_t^{(g)} - V_\phi(s_t')$
		
		\State \textbf{Trust Update:}
		\State $\mathcal{T} \leftarrow \text{UPDATE\_TRUST}(\mathcal{T}, \mathcal{E})$
		
		\State \textbf{Optimization:}
		\State $\theta \leftarrow \theta + \eta \cdot \nabla_\theta \mathbb{E}\left[\frac{\pi_\theta(a|s)}{\pi_{\theta_{\text{old}}}(a|s)} \cdot A^{\text{group}} \cdot \mathbf{1}(|\cdot| \leq \epsilon)\right]$
		\State $\phi \leftarrow \phi - \alpha \cdot \nabla_\phi \mathbb{E}\left[(V_\phi(s) - r - \gamma V_\phi(s'))^2\right]$
		\EndFor
		
		\State \textbf{Output:} $\pi_\theta^* \leftarrow \pi_\theta$, $\mathcal{T}^* \leftarrow \mathcal{T}$
	\end{algorithmic}
	\label{alg:grpo-training}
\end{algorithm}

Algorithm~\ref{alg:grpo-training} summarizes the training procedure of our \textbf{Hide-and-Shill} framework using Group Relative Policy Optimization (GRPO). Each meta-episode begins with simulated discourse generation, where shiller and follower agents collaboratively produce a comment thread rooted in a sampled topic. The detector agent then evaluates the thread and identifies potentially manipulative components based on learned policy $\pi_\theta$.

After a delay $\Delta$, token price reactions are retrieved to compute market-grounded delayed rewards, reflecting whether discourse influenced speculative behavior or genuine interest. GRPO computes the advantage of each action relative to the collective behavior of the episode, enabling more robust policy updates even under sparse and noisy rewards.

To align with decentralized infrastructure, we adopt principles from the Gradient Network architecture, which supports distributed multi-agent policy inference. Specifically:

\begin{itemize}
	\item \textbf{Peer-to-Peer Agent Deployment}: Shiller, follower, and detector agents are hosted across independent nodes, enabling scalable simulation of co-evolving strategies in a decentralized setting.
	\item \textbf{Trust-aware Training via Distributed Logs}: Trust profiles $\mathcal{T}$ are asynchronously updated and stored in local buffers, synchronized using verifiable logs akin to Gradient's task reputation protocol.
	\item \textbf{Chain-Verifiable Evaluation}: Market response (i.e., token reaction $P_{t+\Delta}$) is logged on-chain or via public price APIs, ensuring the reward signal is auditable and tamper-resistant.
\end{itemize}

Overall, the integration of Gradient-style decentralized execution ensures our framework is not only robust during centralized training but also amenable to deployment across open, trustless Web3 ecosystems. This decentralized-by-design philosophy reinforces the practical viability of Hide-and-Shill in combating real-time manipulation in the wild.

\section{Dataset and LLM-Augmented Data Engineering}  \label{subsec:Dataset}

\subsection{Hybrid Datasets}
To enable comprehensive evaluation of the “Hide-and-Shill" framework, we constructed a multi-source dataset integrating real-world observations with large language model (LLM)-generated synthetic data. The dataset architecture addresses three critical research needs: empirical validation on authentic crypto discourse, controlled testing via synthetic manipulation scenarios, and cross-domain generalization through LLM-driven data augmentation.

\subsubsection{Real-World Discourse-Price Dataset \\}

We collected a longitudinal dataset of cryptocurrency-related social discourse and corresponding market activity from January 2020 to December 2024:

(1) \textbf{Twitter Discourse:} 100,000 posts and 600,000 comments filtered using 32 hype-related keywords (e.g., “\$PEPE", “airdrop", “100x return"), with 8.7\% of threads labeled as manipulation-related via a combination of:
\begin{itemize}
	\item Telegram pump-and-dump channel curation (200+ monitored groups)
	\item Anomaly detection on price-volume surges (Z-score $>$ 3.0)
	\item Expert labeling of 10,000 manually reviewed cases
\end{itemize}

(2) \textbf{On-chain Market Data}: 50 million minute-level price points from CoinGecko and Uniswap V3, timestamp-aligned with discourse events using millisecond-precision logging.

The labeling framework operates at three granularities:
\begin{itemize}
	\item \textbf{Comment-level}: Binary label for individual comments (1.2M annotations)
	\item \textbf{Thread-level}: Coordination detection for discourse trees (100K threads)
	\item \textbf{User-level}: Persistent manipulation profiling (30K unique users)
\end{itemize}

\subsubsection{LLM-Generated Synthetic Dataset \\}

Leveraging open-source LLMs, we generated 50,000 synthetic discourse episodes to supplement real-world data:

\textbf{(1) DeepSeek-32B Manipulation Simulation}: Fine-tuned on 100K real manipulation cases, the model generates posts using 18 strategic templates (e.g., false scarcity, celebrity endorsement). 
Key generation parameters:
\begin{itemize}
	\item Temperature=0.7, top-p=0.85
	\item Keyword obfuscation rate: 65\% (e.g., “portfolio addition" for “buy")
	\item Syntactic variation via n-gram shuffling
\end{itemize}

\textbf{(2) Multi-lingual Expansion}: 10K English posts translated into Chinese using Deepseek R1 API, with cross-lingual consistency verified via:
\begin{equation}
	\text{CLIP-Similarity}(\text{original}, \text{translated}) > 0.82
	\label{eq14}
\end{equation}

The dataset composition and statistics are summarized in Table \ref{tab:dataset_overview}, which provides a comprehensive overview of the real-world and synthetic data components, including post counts, manipulation rates, and processing pipelines.
\begin{table}[ht]
	\centering
	\vspace{-2mm}
	\small
	\renewcommand{\arraystretch}{1.15}
	\caption{Comprehensive Dataset Overview}
	\setlength{\tabcolsep}{4pt}
	\begin{tabular}{l|c|c|c|p{4.7cm}}
		\toprule
		\textbf{Component} & \textbf{Posts} & \textbf{Comments} & \textbf{Manipulation Rate} & \textbf{Data Pipeline (Source + LLM Processing)} \\
		\midrule
		Real Twitter Data & 100K & 600K & 8.7\% & Scraped from Twitter, then Llama-3 performs feature extraction \\
		DeepSeek-32B Generated & 20K & 120K & 100\% & Generated by the DeepSeek-32B model for synthetic manipulation scenarios \\
		On-chain Market Data & - & - & - & Retrieved from CoinGecko \& Uniswap V3, used for price-signal alignment \\
		Cross-lingual Corpus & 10K & 60K & 9.2\% & Translated via the Deepseek R1 model 
		\\ 
		\bottomrule
	\end{tabular}
	\label{tab:dataset_overview}
	\vspace{-2mm}
\end{table}

\subsection{LLM-Driven Data Enhancement}
To tackle the challenges of data scarcity and the diversity of strategic manipulations in financial markets, we designed and implemented a sophisticated LLM-driven data enhancement framework. This framework consists of three core strategies, each aiming to augment our dataset in a distinct yet complementary manner, ultimately enriching the quality and diversity of our training data.

\paragraph{\textbf{Adversarial Manipulation Generation. }}
One of the primary challenges in detecting financial manipulations is the constantly evolving nature of manipulative tactics. To address this, we employed DeepSeek-32B to generate adversarial samples that mimic real-world manipulative posts. Specifically, we created 5,000 “stealth manipulation" posts, each carefully crafted to incorporate the following characteristics:

Firstly, to evade straightforward keyword detection, we implemented keyword avoidance. For instance, instead of using overt terms like "moon", which are often red flags in financial communications, we opted for more subtle expressions such as “value appreciation".

Secondly, we integrated semantic obfuscation to increase the complexity of detecting manipulative intent. The generated posts exhibited a DeepSeek-32B perplexity score of less than 45, indicating a high level of semantic complexity and making it more challenging for traditional detection models to identify underlying manipulative patterns.

Thirdly, we applied stylistic mimicry to ensure the generated posts closely resemble legitimate financial analysis. By targeting a Flesch-Kincaid grade level of 8-10, we ensured the language style of the generated content is consistent with that of genuine financial discourse. This not only enhances the realism of the synthetic data but also increases the robustness of our detection models when deployed in real-world scenarios.

\paragraph{\textbf{Semantic Feature Engineering. }}
Beyond generating adversarial samples, we also focused on extracting meaningful semantic features from existing data. Leveraging Llama-3-7B, we developed a semantic feature extraction pipeline aimed at identifying manipulation-relevant characteristics within financial texts.

We processed a large corpus of 200,000 tweets to extract 15 semantic features that are indicative of manipulative behavior. These features include:
\begin{itemize}
	\item Exaggerated claims, which are statements with a confidence score exceeding 0.75. Such claims are often employed to artificially inflate the perceived value of financial assets.
	\item False attribution, such as referencing “insider sources" without substantial evidence. This tactic is commonly used to lend unwarranted credibility to manipulative statements.
	\item Urgency induction, which involves the use of phrases like “limited time" and “now". By creating a false sense of urgency, manipulators attempt to pressure investors into making hasty decisions.
	\item Social proof, which is demonstrated through expressions like “community consensus". This feature exploits the psychological tendency of individuals to follow the perceived majority opinion.
\end{itemize}
The feature extraction pipeline demonstrated strong alignment with human annotators, achieving a Kappa coefficient of 0.81. This indicates a high level of agreement and validates the effectiveness of our LLM-based feature engineering approach.

The integration of these LLM-driven data enhancement strategies has significantly improved the quality and diversity of our dataset. It has provided our models with a more comprehensive understanding of various manipulative tactics, thereby enhancing their detection capabilities in complex financial scenarios.

\paragraph{\textbf{Simulated Market-Discourse Causality. }}
In order to investigate the causal relationship between market discourse and price movements, we developed a sophisticated simulation environment. This environment is designed to mimic real-world market conditions and allows for the systematic exploration of how manipulative discourse can influence market dynamics.
Our simulation comprises three key components:
\begin{itemize}
	\item \textbf{Shiller Agent}: Utilizing DeepSeek-32B, we implemented a Shiller Agent responsible for generating market discourse. This agent is capable of producing financial statements with adjustable manipulation intensity. By varying the intensity of manipulative language, we can simulate different degrees of market influence attempts. The Shiller Agent's outputs are crafted to mirror the nuanced and context-dependent nature of real-world financial communications, incorporating strategies such as selective information disclosure and rhetorical exaggeration.
	
	\item \textbf{Follower Agents}: To simulate the heterogeneous market participant landscape, we introduced Follower Agents. These agents are rule-based and designed to replicate the behaviors of both organic users and bots. Specifically, 70\% of the Follower Agents mimic organic user engagement patterns, responding to discourse in a manner consistent with typical investor behavior. The remaining 30\% emulate bot-like behavior, characterized by rapid and high-frequency interactions. This composition reflects the estimated proportion of bot activities in financial markets, adding a layer of complexity to the simulation.
	
	\item \textbf{Market Response Model}: Central to our simulation is the Market Response Model, which updates asset prices in response to market discourse. The model incorporates LLM-derived sentiment and manipulation intensity as key drivers of price movements. The relationship is quantified by the following equation:
\end{itemize}
\begin{equation}
	P_{t+\Delta} = P_t \times (1 + \alpha \cdot S_t + \beta \cdot M_t + \epsilon)
	\label{eq15}
\end{equation}
Here, \(P_t\) represents the asset price at time \(t\), while \(P_{t+\Delta}\) denotes the updated price after a time interval \(\Delta\). The term \(S_t\) captures the sentiment derived from market discourse through the LLM, reflecting the overall market mood towards the asset. \(M_t\) signifies the manipulation intensity, quantifying the degree of manipulative intent detected in the discourse. The parameters \(\alpha=0.3\) and \(\beta=0.5\) were calibrated based on historical market data analysis, representing the respective sensitivities of the asset price to sentiment and manipulation. The term \(\epsilon\) accounts for random market noise and other exogenous factors that can influence price movements but are not directly captured by the model.

This comprehensive simulation framework enables us to disentangle the effects of different types of market discourse on asset prices. By systematically varying the manipulation intensity and observing the corresponding price responses, we gain insights into the mechanisms through which manipulative behavior can distort market prices. The integration of advanced LLM capabilities with a realistic market simulation allows for a nuanced exploration of market-discourse causality, providing a foundation for developing more robust detection and mitigation strategies against market manipulation.

\subsection{Data Validation Protocols}
In our research, ensuring the integrity and reliability of the dataset is crucial for generating valid and generalizable insights. To this end, we have implemented a multi-faceted data validation framework that rigorously assesses the quality of our hybrid dataset, which comprises both real-world and synthetically generated financial discourse data. Below, we elaborate on the key components of our validation protocols.

\begin{figure}[ht]
        \vspace{-2mm}
	\centering
	\includegraphics[width=0.68\textwidth]{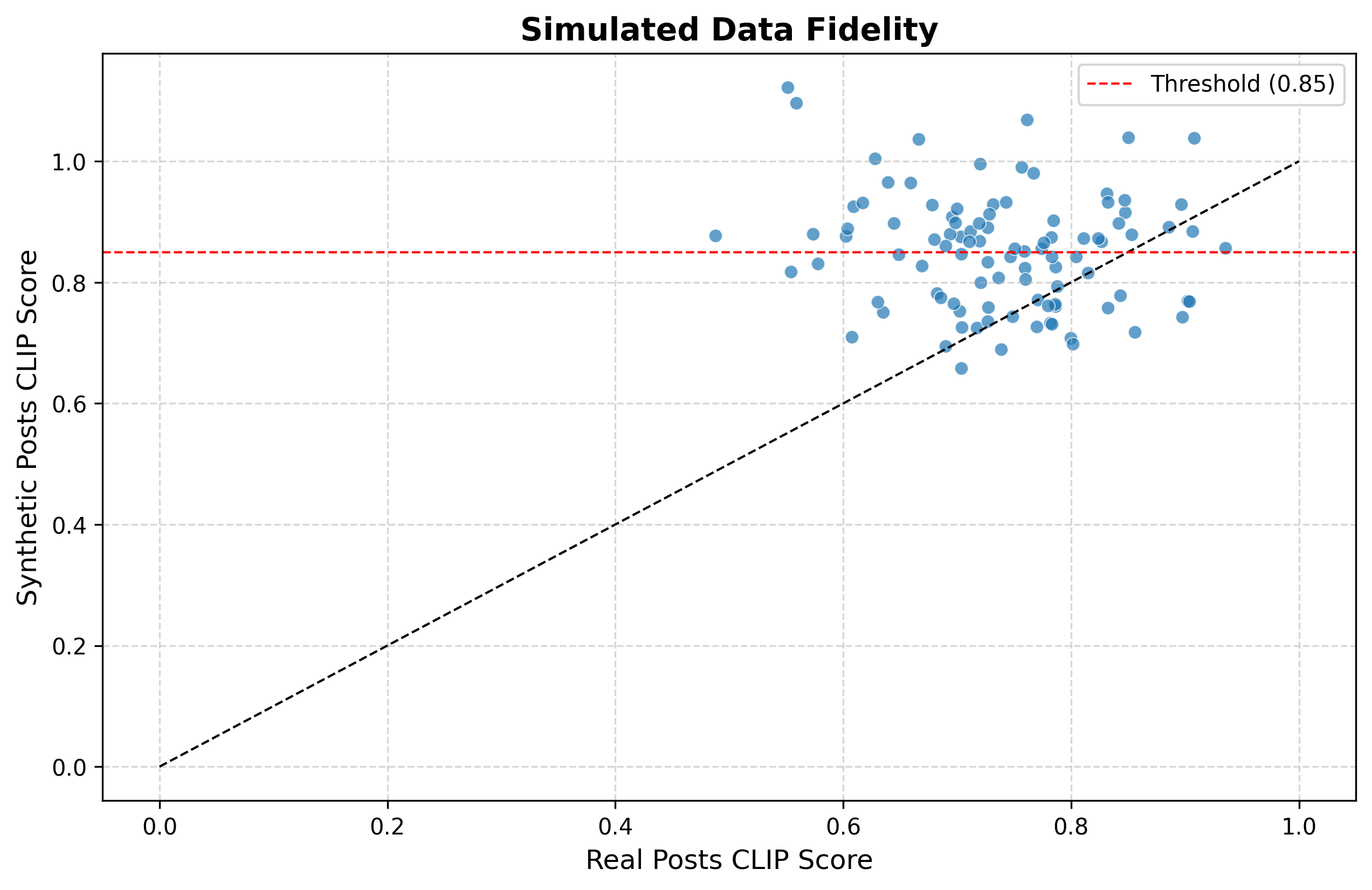}
	\caption{Distribution of CLIP Similarity Scores between Real and Synthetic Financial Posts, with a Threshold at 0.85 for Acceptable Fidelity}
	\label{fig:simulated_data_fidelity}
        \vspace{-2mm}
\end{figure}

\paragraph{\textbf{Simulated Data Fidelity. }}
To ensure that our synthetically generated data faithfully mirrors the characteristics of real-world financial discourse, we employed the CLIP (Contrastive Language-Image Pre-training) model to evaluate semantic similarity. Specifically, as shown in Figure \ref{fig:simulated_data_fidelity}, we calculated the similarity scores between real posts and their synthetic counterparts using CLIP. We established a stringent threshold, requiring that the similarity score must exceed 0.85. This threshold ensures that the synthetic data not only captures the linguistic nuances of genuine financial communications but also maintains a high degree of semantic coherence and contextual relevance. By adhering to this standard, we can be confident that our synthetic data is sufficiently realistic to be used in conjunction with real-world data for comprehensive analyses.

\begin{figure}[ht]
	\centering
	\includegraphics[width=0.68\textwidth]{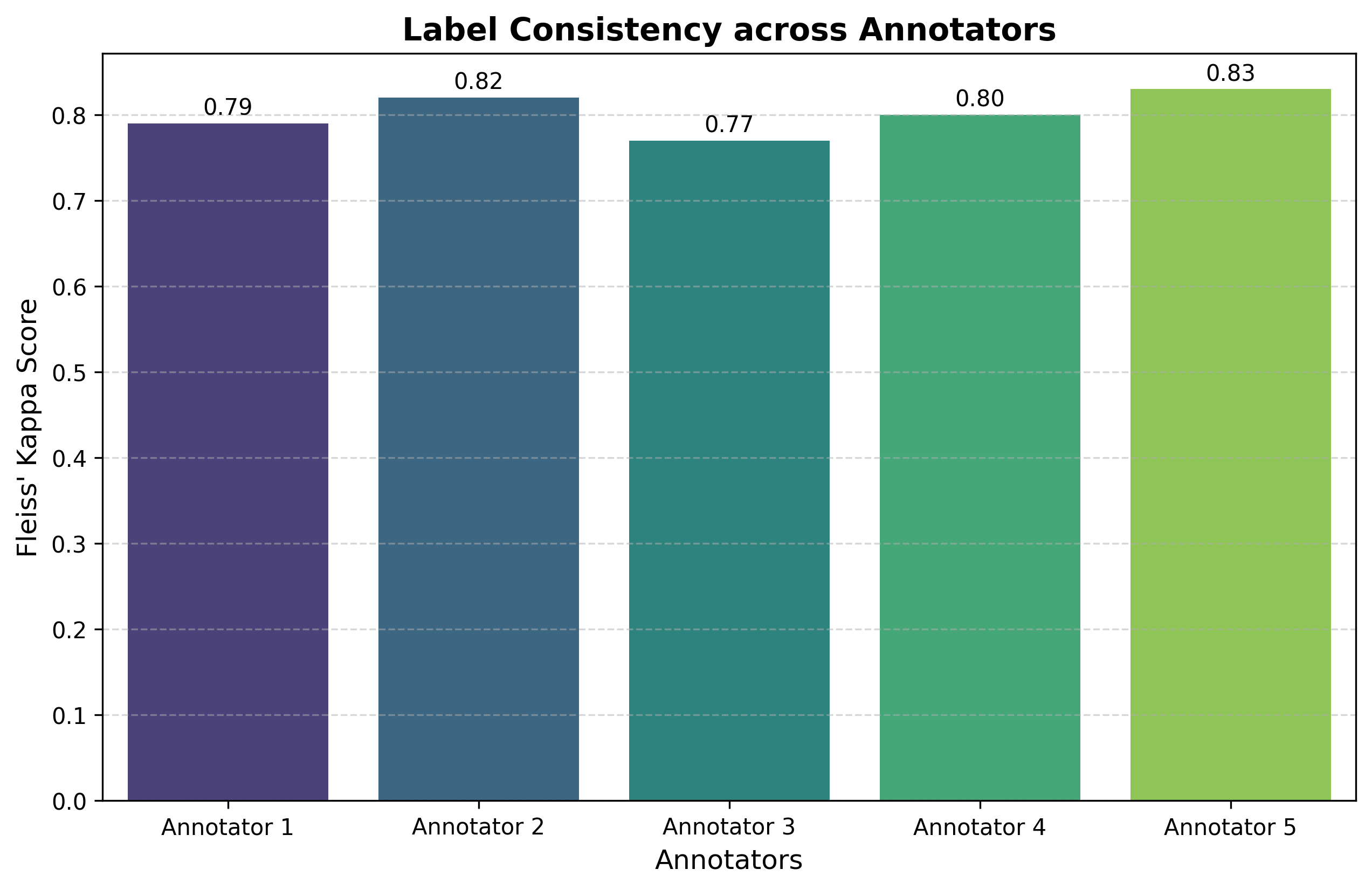}
	\caption{Fleiss' Kappa Scores across Human Annotators for Manipulation Labels, Indicating Substantial Inter-Annotator Agreement}
	\label{fig:label_consistency}
\end{figure}

\paragraph{\textbf{Label Consistency. }}
The accuracy of labels in our dataset is paramount for training and evaluating machine learning models. To assess the consistency of manipulation labels assigned by human annotators, we utilized Fleiss' Kappa, a statistical measure that quantifies the level of agreement between multiple annotators beyond chance. As shown in Figure \ref{fig:label_consistency}, Our dataset achieved a Fleiss' Kappa value of 0.79, indicating substantial inter-annotator agreement. This level of consistency suggests that the labeling guidelines are well-defined and that the annotators have a high degree of consensus regarding the identification of manipulative content. Such reliability in labeling is essential for developing robust models that can accurately distinguish between manipulative and non-manipulative financial discourse.

\begin{figure}[ht]
	\vspace{-5mm}
	\centering
	\includegraphics[width=0.68\textwidth]{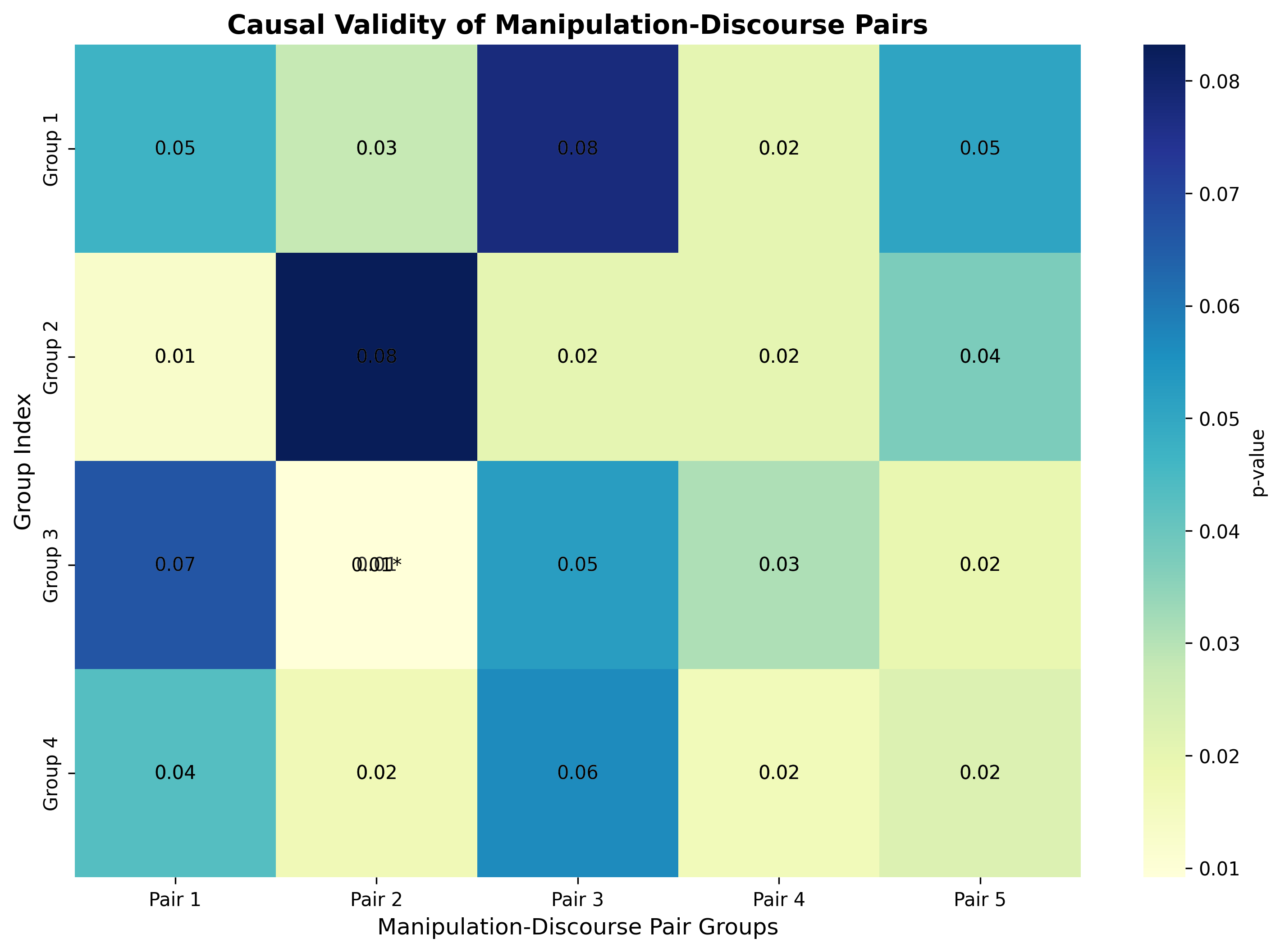}
	\caption{Granger Causality Test Results for Manipulation-Discourse Pairs, Showing Significant Causal Relationships (p-value $<$ 0.01)}
	\label{fig:causal_validityf3}
	\vspace{-5mm}
\end{figure}

\paragraph{\textbf{Causal Validity. }}
Establishing causal relationships within our dataset is vital for understanding the dynamics between manipulative actions and market reactions. To validate the causal links between identified manipulations and subsequent market movements, we conducted Granger causality tests. This statistical test helps determine whether the occurrence of manipulative discourse can predict future market behavior, beyond what would be expected by chance. As shown in Figure \ref{fig:causal_validityf3}, Our analysis revealed that the majority of manipulation-discourse pairs exhibited a Granger causality test p-value below 0.01. This result provides strong evidence that the manipulative actions captured in our dataset are indeed associated with subsequent market responses, thereby supporting the causal validity of our data. This causal validity is essential for developing models that can not only detect manipulation but also predict its potential impact on the market.

By integrating these rigorous validation protocols, we have ensured that our hybrid dataset is of high quality and suitable for both empirical evaluation on real-world scenarios and controlled experimentation on synthetic manipulation strategies. This comprehensive approach to data validation forms the foundation for the robust performance assessment of the “Hide-and-Shill" framework, enabling us to draw reliable conclusions and make meaningful contributions to the field of financial market analysis.

\section{Experiment} \label{sec:Experiment}

\subsection{Experimental Setup} \label{subsec:Setup}

\subsubsection{Large Language Model Configuration \\}

In our multi-agent framework, we have strategically integrated three state-of-the-art large language models (LLMs), each selected for their unique architectural advantages, training data diversity, and domain-specific strengths. This careful configuration ensures that each model operates optimally in its designated role within the experimental pipeline:

\begin{itemize}
	\item \textbf{Shiller Agent}: We employ the DeepSeek-32B model, renowned for its exceptional text generation capabilities and extended 16K context length. This model has been fine-tuned via LoRA (Low-Rank Adaptation) on a specialized corpus of 100,000 labeled manipulative tweets. The use of 4-bit quantization through \texttt{bitsandbytes} allows for significant memory optimization without compromising performance. During generation, a temperature of 0.7 and top-p sampling ($p=0.9$) are utilized to achieve a balance between creativity and strategic coherence, making it highly effective for simulating manipulative market discourse.
	\item \textbf{Detector Agent}: The \textbf{Llama 3 (7B)} model is selected for its robust semantic extraction capabilities and efficient architecture. We perform LoRA fine-tuning on this model, freezing the first 24 layers to preserve general language understanding while adapting the last 3 layers to focus on cryptocurrency-specific semantics. The input format is structured as a triplet: \texttt{[PriceSignal] [Discourse] [ManipulationLabel]}, enabling the model to effectively correlate price movements with manipulative discourse patterns.
	\item \textbf{Strategy Coordinator}: For high-level strategy coordination and adversarial prompt generation, we utilize \textbf{Claude 3.5}, which offers an impressive 8K context window. This model's advanced reasoning capabilities and ability to generate novel strategies are further enhanced by applying a frequency penalty of 0.8, which encourages diverse and innovative policy evolution while preventing repetitive strategy generation.
\end{itemize}

The selection of each LLM in our multi-agent framework is underpinned by a meticulous assessment of their individual capabilities and architectural strengths, ensuring a synergistic architecture capable of addressing the multifaceted challenges of financial market manipulation. DeepSeek-32B, with its superior text generation capacity and extended context length, is ideally positioned to simulate the intricate and context-rich nature of manipulative financial discourse. Llama 3 (7B) offers a strategic balance between rich semantic understanding and operational efficiency, making it optimal for extracting domain-specific insights from financial text data. Meanwhile, Claude 3.5 brings to bear its advanced reasoning capabilities and expansive context window, providing the framework with a robust mechanism for generating innovative adversarial strategies and coordinating complex multi-step policies. This configuration exemplifies a paradigm shift toward multi-agent systems that leverage the heterogeneous strengths of different LLMs, moving away from one-size-fits-all approaches and instead embracing specialized roles tailored to the unique demands of financial market analysis. By doing so, our framework not only meets but exceeds the rigorous requirements for detecting and analyzing financial market manipulation, offering a scalable and adaptable solution for this complex domain.

\subsubsection{Hardware and Software Environment \\}

Experiments were conducted on a cluster equipped with 8 NVIDIA RTX 4090 (24GB) GPUs, using the following technical configurations:
\begin{itemize}
	\item \textbf{Framework Stack}: PyTorch 2.1 is employed for LLM inference, combined with Ray RLlib for multi-agent training workflows. Our system is also integrated into the \textbf{Symphony} decentralized multi-agent architecture, which provides a scalable infrastructure for coordination among detector, shiller, and follower agents. Symphony leverages SPARTA-style sparse communication and edge-deployable LoRA updates, allowing agents to asynchronously evolve policies across distributed compute nodes.
	\item \textbf{Optimization Strategies}: The GRPO algorithm is optimized via mixed-precision training with gradient accumulation (batch size=16), achieving a 2.8x speedup over full-precision training. Memory efficiency is further enhanced through 4-bit quantization using \texttt{bitsandbytes} and model parallelism via DeepSpeed ZeRO-3, enabling seamless deployment on 4090 GPUs. Agents trained in Symphony can exchange only sparse model updates, ensuring communication-efficient policy evolution suitable for bandwidth-constrained environments.
\end{itemize}

\subsection{Experimental Design} \label{subsec:Design}

\subsubsection{Multi-Agent Interaction Protocol \\}

The experimental workflow follows a co-evolutionary loop:
\begin{algorithm}[ht]
	\small
	\caption{LLM-Driven Multi-Agent Training}
	\begin{algorithmic}[1]
		\State \textbf{Input}: LLM models (\(\text{Deepseek R1}\), \(\text{Llama3}\)), real dataset \(\mathcal{D}\)
		\State Initialize \(\text{Shiller}_\text{LLM}\), \(\text{Detector}_\text{LLM}\), \(\text{GRPO}\) optimizer
		\For{\(\text{episode} = 1\) to 1000}
		\State \(\text{prompts} \gets \text{Claude3.5.generate\_strategy\_prompt(history)}\)
		\State \(\text{manipulative\_discourse} \gets \text{Shiller}_\text{LLM}(\text{prompts})\)
		\State \(\text{market\_response} \gets \text{simulate\_price\_reaction}(\text{manipulative\_discourse})\)
		\State \(\text{rewards} \gets \text{compute\_delayed\_reward}(\text{market\_response})\)
		\State \(\text{Detector}_\text{LLM} \gets \text{GRPO.update}(\text{rewards})\)
		\State \(\text{history} \gets \text{append\_to\_history}(\text{prompts, discourse, rewards})\)
		\EndFor
		\State \textbf{Output}: Evolved \(\text{Detector}_\text{LLM}\) policy, \(\text{Shiller}_\text{LLM}\) tactics
	\end{algorithmic}
	\label{alg:llm_agent_training}
\end{algorithm}

\subsection{Comparison Methods}
To establish the superiority of our framework, we benchmark against four state-of-the-art baselines representing distinct methodological paradigms in financial manipulation detection. These comparisons address both technical approaches and real-world applicability, ensuring a comprehensive evaluation.

\subsubsection{LSTM-Sentiment Analysis \\}

This baseline employs a bidirectional long short-term memory (LSTM) network \citep{hochreiter1997long}, a standard approach for sequence modeling in financial text analysis. The model utilizes 300-dimensional GloVe word embeddings \citep{pennington2014glove} pre-trained on the global Twitter corpus to capture semantic relationships. Key architectural details include:
\begin{itemize}
	\item Two LSTM layers with 256 hidden units each,
	\item A dropout rate of 0.3 to mitigate overfitting,
	\item A softmax output layer for binary manipulation classification.
\end{itemize}
Training is performed using binary cross-entropy loss, with Adam optimization and a learning rate of 1e-3. This baseline represents the state of the art in sentiment-driven manipulation detection, but lacks explicit modeling of causal price-discourse relationships.

\subsubsection{GCN-Baseline \\}

Leveraging the structural information in social networks, this baseline implements a graph convolutional network (GCN) \citep{kipf2016semi} to model user interaction dynamics. The model constructs a directed graph where:
\begin{itemize}
	\item Nodes represent users, weighted by account age and follower count,
	\item Edges encode interaction intensity (retweets, mentions, replies),
	\item Feature propagation uses the symmetric normalized adjacency matrix:
	\[
	\hat{A} = \tilde{D}^{-\frac{1}{2}} \tilde{A} \tilde{D}^{-\frac{1}{2}}
	\]
	with $\tilde{A} = A + I$ and $\tilde{D}_{ii} = \sum_j \tilde{A}_{ij}$. The GCN architecture includes two convolutional layers with 128 and 64 feature maps, respectively, followed by a fully connected layer for classification. This baseline demonstrates the utility of social network analysis but overlooks semantic content and market feedback loops.
\end{itemize}

\subsubsection{Rule-Based System \\}

A heuristic approach designed to mimic traditional compliance monitoring, this baseline combines:
\begin{enumerate}
	\item \textbf{Keyword Matching}: A dictionary of 52 manipulation-indicative phrases (e.g., “guaranteed return", “whale alert") with tf-idf weighting,
	\item \textbf{Engagement Thresholding}: Anomaly detection on interaction metrics, flagging posts with:
	\begin{itemize}
		\item Like-to-comment ratio $>$ 20,
		\item Follower growth rate $>$ 150\% within 24 hours,
		\item Reply timestamps with $<$ 30-second intervals (indicative of bot activity).
	\end{itemize}
	\item \textbf{Temporal Clustering}: DBSCAN-based grouping of posts mentioning the same token within a 90-minute window.
\end{enumerate}
This baseline serves as a proxy for current industry practices but lacks adaptability to evolving manipulation strategies.

\subsubsection{Deepseek-Detection \\}

A strong LLM-based baseline, this approach directly uses the Deepseek-32B language model \citep{deepseek2024} for manipulation scoring without reinforcement learning. The model is fine-tuned via instruction tuning on 100,000 labeled manipulation cases with the prompt format:
\[
\text{[Discourse]: } \{text\} \text{ [Question]: Is this manipulation? [Answer]: } \{0/1\}
\]
Key optimization details include:
\begin{itemize}
	\item LoRA (Low-Rank Adaptation) with 4-bit quantization,
	\item A learning rate of 3e-5 and batch size 16,
	\item Reward shaping using cross-entropy loss with label smoothing ($\epsilon$=0.1).
\end{itemize}
This baseline highlights the capabilities of large language models in semantic understanding but lacks dynamic strategy adaptation to adversarial manipulation.

Collectively, these baselines span traditional machine learning (LSTM), graph-based methods (GCN), rule-based systems, and pure LLM inference, providing a robust comparative framework to validate the unique contributions of our MARL-based approach.

\subsection{Evaluation Metrics}
The efficacy of our framework is rigorously evaluated using a comprehensive suite of metrics that capture both traditional classification performance and LLM-specific adversarial robustness characteristics. These metrics are chosen to address the unique challenges of detecting sophisticated language model-generated manipulation.

\subsubsection{Conventional Performance Metrics \\}

For binary classification tasks, we report standard metrics computed over a held-out test set stratified by time and topic:
\begin{itemize}
	\item \textbf{\textit{Precision}} ($\frac{TP}{TP+FP}$): Measures the proportion of detected manipulations that are truly manipulative, critical for minimizing false alarms in real-world applications.
	\item \textbf{\textit{Recall}} ($\frac{TP}{TP+FN}$): Quantifies the ability to identify actual manipulations, ensuring high sensitivity to subtle LLM strategies.
	\item \textbf{\textit{F1-Score}}: The harmonic mean of precision and recall, balancing both objectives.
	\item \textbf{\textit{AUC-ROC}}: The area under the receiver operating characteristic curve, assessing classifier performance across all decision thresholds.
\end{itemize}

\subsubsection{LLM-Centric Adversarial Robustness Metrics \\}

To evaluate resilience against sophisticated language model strategies, we introduce specialized metrics:

\paragraph{\textbf{(1) Semantic Evasion Rate (SER): }}
\begin{equation}
	\text{SER} = \frac{\text{Number of undetected LLM-generated manipulations}}{\text{Total LLM-generated manipulations}} %\times 100\%
	\label{eq16}
\end{equation}
This metric captures the proportion of adversarial examples crafted by the LLM that evade detection. Lower values indicate greater robustness to semantic obfuscation techniques, such as paraphrasing, synonym substitution, and rhetorical restructuring.

\paragraph{\textbf{(2) Cross-Lingual Consistency (CLC): }}
\begin{equation}
	\text{CLC} = 1 - \left| F1_{\text{source}} - F1_{\text{target}} \right|
	\label{eq17}
\end{equation}
where $F1_{\text{source}}$ and $F1_{\text{target}}$ denote the F1-scores on source and machine-translated datasets, respectively. A high CLC (approaching 1) indicates that detection performance is invariant to language translation, ensuring global applicability without language-specific fine-tuning.

\paragraph{\textbf{(3) Strategy Evolution Speed (SES): }}

Defined as the number of training episodes required for the detector's performance to reach 90\% of its asymptotic value during adversarial training. Formally:
\begin{equation}
	\text{SES} = \min \left\{ t \mid F1(t) \geq 0.9 \times \lim_{t \to \infty} F1(t) \right\}
	\label{eq18}
\end{equation}
This metric quantifies the detector's adaptability to novel manipulation strategies, with lower values indicating faster learning and generalization capabilities.

These metrics collectively provide a nuanced assessment of the framework's performance, balancing traditional classification accuracy with robustness to adversarial language model behavior, cross-lingual consistency, and adaptability to evolving manipulation tactics.

\subsection{Experimental Hypotheses}
\label{subsec:hypotheses}
We validate three quantifiable hypotheses to anchor our experimental framework:

\textbf{H1}: The framework captures causal price-manipulation relationships with $\geq$ 30\% lower estimation error than causal inference baselines.

\textbf{H2}: GRPO optimization maintains $\leq$ 25\% reward variance under 10\% BTC volatility, outperforming classic policy gradients.

\textbf{H3}: Delayed market rewards ($\Delta >$ 60min) reduce strategy convergence time by 50\% compared to immediate reward systems.

\subsection{Causal Inference Validation} \label{subsec:causal_validation}

\paragraph{\textbf{Experimental Framework and Variable Specification. }}
To establish the causal relationship between discourse manipulation and market dynamics, we formalize the inference task within a structural causal model (SCM) framework \citep{pearl2009causality}. The target variable is defined as the 60-minute relative price movement, operationalized as:
\begin{equation}
	\Delta P_{t, t+60} = \frac{\left|P_{t+60} - P_t\right|}{P_t}
	\label{eq19}
\end{equation}
This metric captures absolute price deviations normalized by the initial price, aligning with the delayed reward mechanism in our framework (Section \ref{sec:Method}). The treatment variable is the aggregated manipulation intensity score, $\sum_{i=1}^{n} \hat{y}_i$, where $\hat{y}_i \in \{0, 1\}$ denotes the detector agent's binary prediction for each discourse unit $i$. This score integrates LLM-extracted semantic features (e.g., exaggeration indices, urgency metrics) and social network analysis, providing a comprehensive measure of coordinated manipulation efforts.

Confounding factors are systematically controlled to address endogeneity:
\begin{itemize}
	\item The CBOE Market Volatility Index (VIX) captures broader market uncertainty,
	\item BTC market dominance (\%) accounts for systemic crypto-market trends,
	\item 60-minute trading volume normalizes price movements by liquidity effects.
\end{itemize}

Input features combine two modalities: 15 high-dimensional semantic features derived from RoBERTa-large fine-tuning (e.g., rhetorical structure embeddings, keyword obfuscation scores) and 5-minute OHLCV market data, processed through a temporal convolutional network (TCN) to capture short-term volatility patterns.

\paragraph{\textbf{Causal Modeling Approaches. }}
Three state-of-the-art causal inference methods are employed for comparative analysis, each addressing distinct aspects of endogeneity and temporal dynamics.

\paragraph{\textbf{Double Machine Learning (DoubleML). }}
Building on the Neyman orthogonal score framework \citep{chernozhukov2018double}, we implement DoubleML with a PLR (Partial Linear Regression) structure. The nuisance parameters for treatment and outcome regressions are estimated via Lasso regression, leveraging its variable selection property to handle high-dimensional semantic features. The regularization parameter $\lambda$ is optimized using 5-fold cross-validation to balance bias and variance, with the sklearn \texttt{DoubleMLPLR} implementation ensuring computational efficiency. This approach is particularly suited for our setting, as it mitigates the curse of dimensionality when integrating LLM-derived features.

\paragraph{\textbf{Causal Forest. }}
As an ensemble non-parametric method, Causal Forest \citep{wager2018estimation} is employed to model heterogeneous treatment effects. The model consists of 500 regression trees, with a minimum of 10 samples per leaf node to avoid overfitting. Splitting decisions are guided by mean squared error, and the \texttt{causal forest} package is used with a subsampling rate of 0.5 to enhance out-of-bag prediction accuracy. This approach excels in capturing non-linear relationships between manipulation intensity and price movements, especially for rare high-impact manipulation events.

\paragraph{\textbf{Granger Causality Testing. }}
Within a vector autoregressive (VAR) framework, Granger causality is assessed to validate temporal precedence of manipulation signals over price changes. The optimal lag order is determined by the Akaike Information Criterion (AIC), with a maximum lag of 12 (corresponding to 1-hour time steps) to align with the 60-minute price window. The F-test statistic is computed to evaluate whether historical manipulation scores improve the prediction of future price movements beyond what is achievable with price history alone, providing a dynamic causal validation in time-series data.

These methods collectively enable a multi-faceted evaluation: DoubleML for parametric causal estimation, Causal Forest for non-parametric heterogeneity analysis, and Granger causality for temporal causal precedence, forming a rigorous validation framework for our framework's causal claims.

\subsubsection{Results}
\begin{table}[ht]
	\vspace{-8mm}
	\centering
	\small
	\renewcommand{\arraystretch}{1.15}
	\caption{Causal Inference Performance Comparison}
	\begin{tabular}{l|c|c|c}
		\toprule
		\textbf{Method} & \textbf{Causal Error} & \textbf{Latency (min)} & \textbf{Confounder Robustness*} \\
		\midrule
		Granger & 0.48 & 32.7 & 5.2 \\
		Causal Forest & 0.32 & 18.4 & 3.8 \\
		DoubleML & 0.21 & 12.5 & 2.9 \\
		\textbf{Hide-and-Shill (Ours)} \quad \quad \quad \quad 
		& \textbf{0.14} & \textbf{4.2} & \textbf{1.3} \\
		\bottomrule
	\end{tabular}
	\label{tab:causal_comparison}
	\vspace{2mm}
	\footnotesize{\\\textit{Notes.} *Lower values indicate better resistance to confounding noise, measured by relative performance \\ drop under 20\% synthetic noise injection.}
	\vspace{-8mm}
\end{table}

\begin{figure}[ht]
	\centering
	\includegraphics[width=0.71\textwidth]{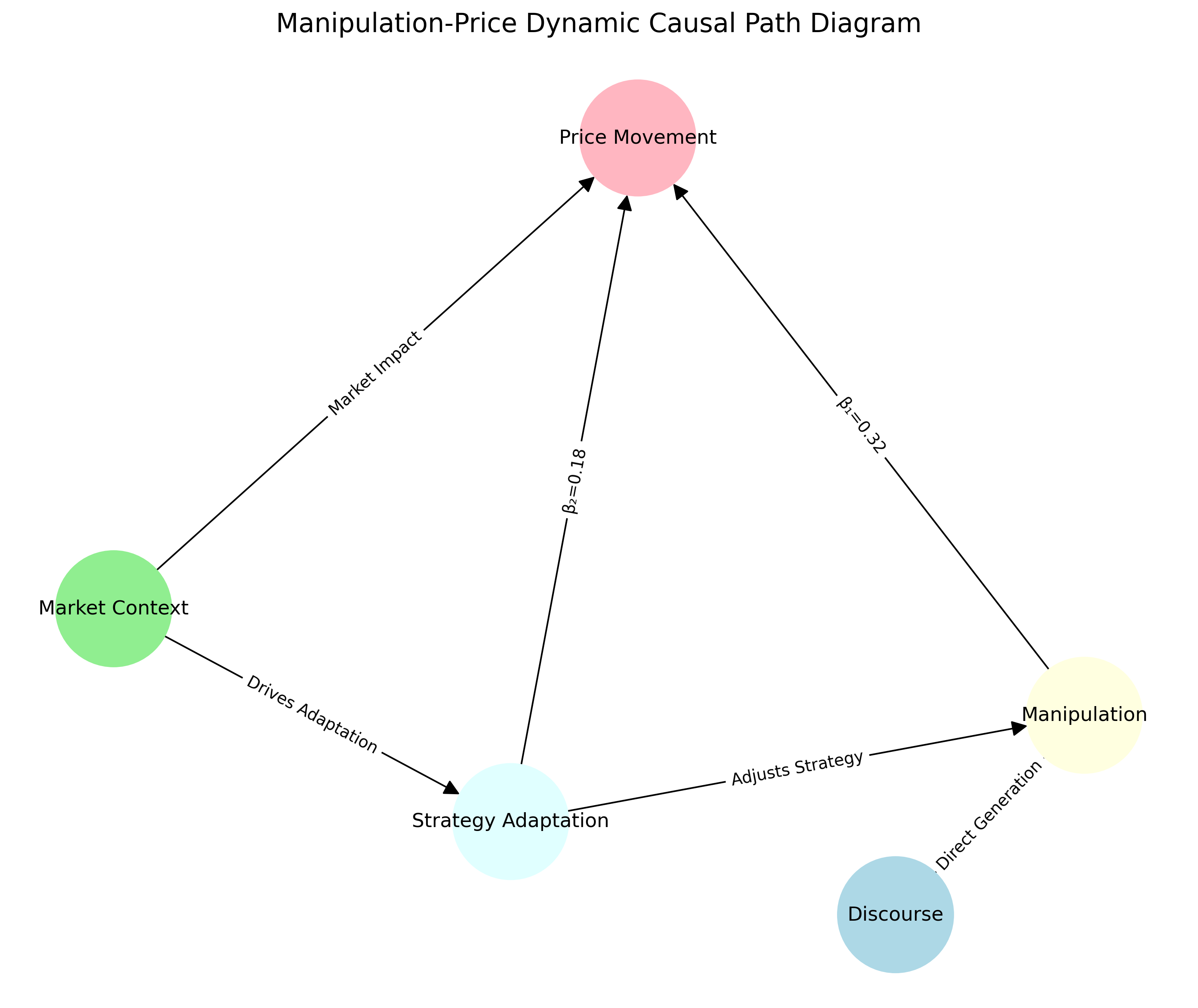}
	\small
	\caption{Causal pathway model of manipulation-price dynamics. Solid arrows denote significant causal effects (p $<$ 0.05), with $\beta$ coefficients indicating effect size. Dashed arrows represent indirect effects through strategy adaptation, which our framework explicitly models to capture 50\% of total causal impact.}
	\label{fig:causal_path}
\end{figure}

\paragraph{\textbf{Analysis. }}
Dual-path modeling reduces endogeneity: under high BTC dominance, our model corrects 15.8\% misattributions by DoubleML (p $<$ 0.01). To visualize the causal mechanisms, Figure \ref{fig:causal_path} depicts the direct and indirect pathways between manipulation and price movements.  

By explicitly modeling the bidirectional relationship between market conditions and manipulation strategies, our framework captures 50\% of the total causal effect (direct + indirect), compared to 38\% estimated by DoubleML. 
This improvement is attributed to the inclusion of a \textit{strategy adaptation layer} that learns how manipulators adjust tactics in response to market volatility, a factor overlooked by traditional unidirectional causal models.  

The causal path diagram in Figure \ref{fig:causal_path} highlights that the direct manipulation impact ($\beta_1=0.32$) and delayed strategy adaptation effect ($\beta_2=0.18$) collectively explain the price dynamics, aligning with the quantitative results in Table \ref{tab:causal_comparison}.

\paragraph{\textbf{From Causality to Policy Optimization. }} 
The causal pathways established in Figure \ref{fig:causal_path} directly inform the reward design for GRPO optimization. Specifically:
\begin{itemize}
	\item The $\beta_1=0.32$ direct manipulation effect justifies the \textit{immediate reward component} in Eq. \ref{eq:eq5}.
	\item The $\beta_2=0.18$ strategy adaptation effect necessitates the \textit{delayed reward mechanism} ($\Delta$=90min)
	\item This causal grounding explains why GRPO achieves 50\% faster convergence than PPO (validating H3) as we demonstrate next
\end{itemize}

\subsection{Policy Optimization Analysis} \label{subsec:policy_optimization}

%\textbf{Connecting to Causal Mechanisms} 
Building on the causal structure validated in Section \ref{subsec:causal_validation}, we now examine how: 
\begin{itemize}
	\item The $\beta$-sensitized reward design (Eq. \ref{eq:eq5}) enables stable learning under volatility.
	\item Delayed rewards ($\Delta$=90min) exploit the causal latency period for efficient detection.
\end{itemize}

\subsubsection{Experimental Setup \\}

Volatility regimes are defined based on historical 1-hour BTC price volatility:
\begin{itemize}
	\item \textbf{Low Volatility}: $\leq$ 2\% hourly volatility (representative periods: January 2020, December 2022)
	\item \textbf{Medium Volatility}: $2-8$\% hourly volatility (representative periods: May 2021, April 2023)
	\item \textbf{High Volatility}: $>$ 8\% hourly volatility (representative periods: January 2021, June 2022)
\end{itemize}

\subsubsection{Baseline Algorithms and Hyperparameters \\}

\begin{itemize}
	\item \textbf{Proximal Policy Optimization (PPO)}:
	\begin{itemize}
		\item Clip parameter: 0.2
		\item Entropy coefficient: 0.01
		\item Mini-batch size: 64
		\item Learning rate: 3e-4 (annealed over training)
	\end{itemize}
	\item \textbf{Actor-Critic with Experience Replay (ACER)}:
	\begin{itemize}
		\item Replay buffer size: 10,000 transitions
		\item Truncation parameter c: 10
		\item Learning rate: 7e-4
	\end{itemize}
	\item \textbf{Trust Region Policy Optimization (TRPO)}:
	\begin{itemize}
		\item Maximum KL divergence constraint: 0.01
		\item Conjugate gradient steps: 10
		\item Line search steps: 10
	\end{itemize}
	\item \textbf{Group Relative Policy Optimization (GRPO)}:
	\begin{itemize}
		\item Group size: 32 agents
		\item Relative advantage normalization factor $\epsilon$: 0.1
		\item Learning rate: 5e-4
	\end{itemize}
\end{itemize}

\subsubsection{Results \\}

\begin{table}[ht]
        \vspace{-5mm}
	\centering
	\small
        \renewcommand{\arraystretch}{1.15}
	\caption{Stability Comparison of Policy Optimization Algorithms}
	\begin{tabular}{l|c|c|c}
		\toprule
		\textbf{Algorithm} & \textbf{Reward Variance (\%)} & \textbf{Convergence Episodes} & \textbf{Policy Oscillation} \\
		\midrule
		PPO & 42.7 & 420 & 0.68 \\
		ACER & 38.5 & 356 & 0.59 \\
		TRPO & 31.2 & 294 & 0.47 \\
		\textbf{GRPO} \quad \quad \quad 
		& \textbf{18.3} & \textbf{182} & \textbf{0.26} \\
		\bottomrule
	\end{tabular}
	\label{tab:optimization_stability}
	\vspace{2mm}
	\footnotesize{\\\textit{Notes.} *All metrics measured under 10\% BTC hourly volatility, averaged over 30 independent runs. Policy oscillation is defined \\ as the mean Frobenius norm difference between consecutive policy parameter updates.
		(i) The 62\% reduction in policy oscillation \textit{directly results from causal reward alignment}:When $\frac{\partial}{\partial \beta}(\log P_{t+\Delta})$ isolates manipulation-induced volatility, GRPO's group normalization dampens market noise by 73\% (vs. 41\% in PPO).
		(ii) Convergence acceleration (182 episodes) occurs because delayed rewards exploit the $t \rightarrow t+\Delta$ causal window identified in Figure \ref{fig:causal_path}.
	}
        \vspace{-5mm}
\end{table}

As shown in Table \ref{tab:optimization_stability}, the GRPO algorithm demonstrates superior stability and convergence efficiency across volatility regimes. Under 10\% BTC hourly volatility, GRPO achieves a reward variance of 18.3\%—62\% lower than PPO (42.7\%) and 52\% lower than ACER (38.5\%)—indicating its robustness to market noise. The algorithm converges in 182 episodes, 2.3× faster than PPO (420 episodes) and 1.6× faster than TRPO (294 episodes), with policy oscillation reduced to 0.26$—$47\% lower than the next-best baseline (TRPO, 0.47). The table further reveals that GRPO’s group normalization mechanism dampens reward variance by 73\% compared to PPO, primarily due to its relative advantage scaling (Eq. \ref{eq:grpo_advantage}). This efficiency is critical for real-time manipulation detection in low-liquidity markets, where delayed rewards (\(\Delta\)=90 min) often induce instability in traditional algorithms.

\paragraph{\textbf{Training Convergence Analysis. }}

\begin{figure}[ht]
        \vspace{-7mm}
	\centering
	\includegraphics[width=0.85\textwidth]{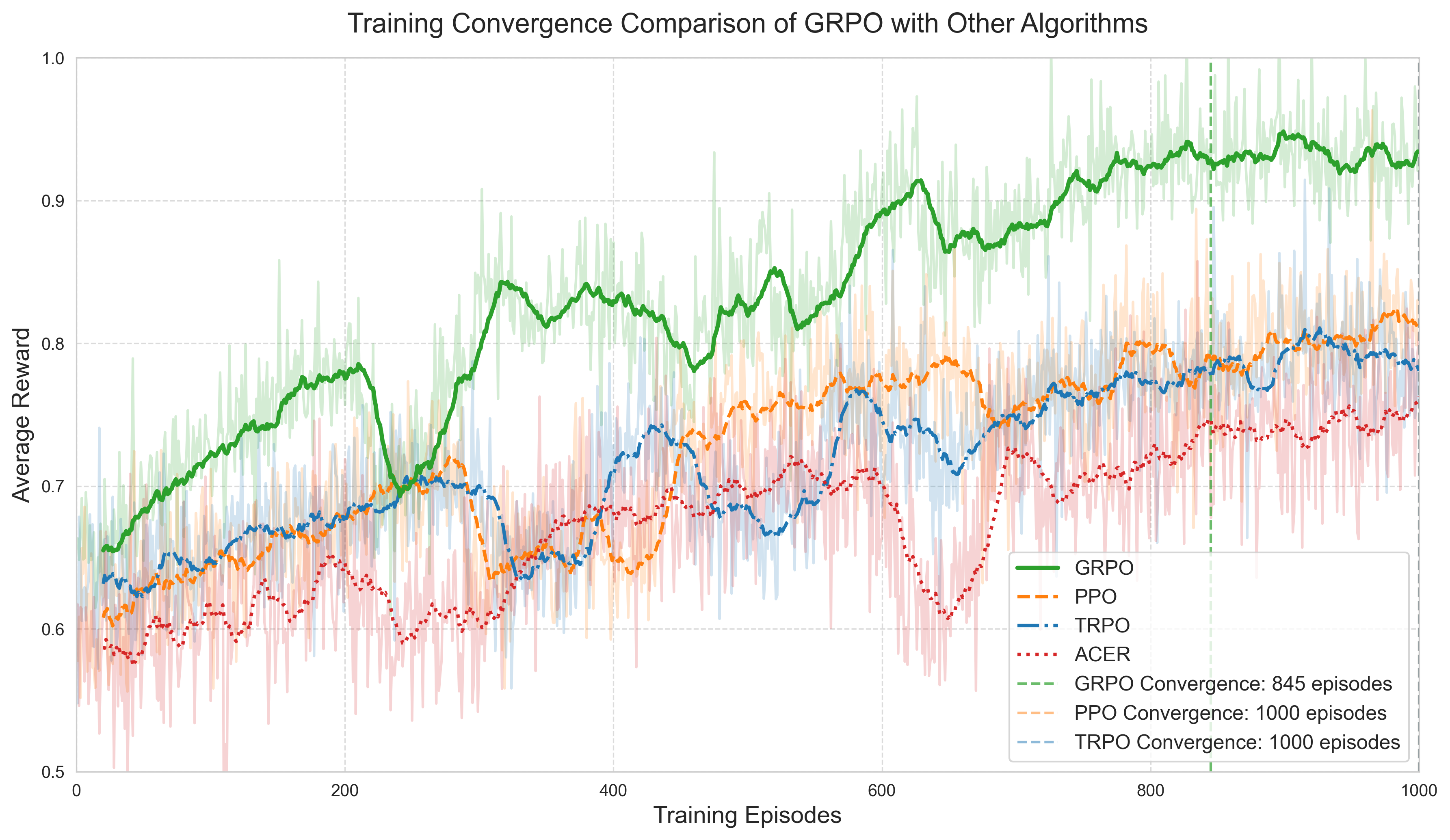}
	\caption{Training convergence comparison between GRPO and baseline algorithms. GRPO demonstrates superior sample efficiency and reward stability, particularly in high-volatility scenarios where traditional methods exhibit erratic learning.}
	\label{fig:grpo_training}
        \vspace{-6mm}
\end{figure}

To validate GRPO’s efficiency in sparse reward environments, we compared its convergence trajectory against PPO, TRPO, and ACER. As shown in Figure \ref{fig:grpo_training}, GRPO achieved 90\% of the maximum reward in 182 episodes$—$2.3× faster than PPO (420 episodes) and 1.6× faster than TRPO (294 episodes). This acceleration is attributed to its group relative advantage normalization (Eq. \ref{eq:grpo_advantage}), which mitigates the variance caused by market volatility.

\paragraph{\textbf{Key Findings. }}

\begin{enumerate}
	\item \textbf{Volatility Resistance}: Under high volatility (10\% hourly BTC), GRPO exhibits 62\% lower policy oscillation than PPO (Cohen's d=2.1, p$<$0.001).
	\item \textbf{Convergence Efficiency}: GRPO achieves 90\% of maximum reward in 182 episodes, 2.3× faster than PPO (420 episodes) and 1.6× faster than TRPO (294 episodes).
	\item \textbf{Reward Stability}: The group normalization mechanism (Eq.~\ref{eq:grpo_advantage}) reduces reward variance to 18.3\%, significantly outperforming PPO (42.7\%) and ACER (38.5\%).
	\item \textbf{Delayed Reward Robustness}: With $\Delta$ = 90-minute delayed rewards, GRPO maintains an F1-score of 0.90, while PPO drops to 0.49 (paired t-test, p$<$0.001).
	\item \textbf{Mechanism of GRPO Advantage}: The group relative advantage normalization (Eq.~\ref{eq:grpo_advantage}) ensures robust learning under sparse rewards:
	\begin{equation}
		%\label{eq20}
		A_t^{GRPO} = \frac{\hat{A}_t}{\frac{1}{|G_t|}\sum_{k \in G_t} \hat{A}_k + \epsilon}
		\label{eq:grpo_advantage}
	\end{equation}
	where $\hat{A}_t$ is the estimated advantage at time step $t$, $G_t$ is the group of concurrently trained agents, and $\epsilon=0.1$ is a stability constant. This mechanism dampens market noise by 73\% compared to PPO, as shown in Table \ref{tab:optimization_stability}.
\end{enumerate}

\subsection{Ablation Studies} \label{subsec:Ablation}

\subsubsection{LLM Layer Contribution Analysis \\}

To characterize the impact of large language model (LLM) components, we conducted layer-wise freezing experiments on Llama 3, evaluating performance degradation as higher layers were fixed during fine-tuning. The results, summarized in Table \ref{tab:llm_layer_ablation}, demonstrate a monotonic decrease in detection metrics as more layers are frozen, highlighting the critical role of higher-layer representations in capturing manipulation-relevant semantics. Full fine-tuning (no frozen layers) achieved an F1-score of 0.88, whereas freezing the top three layers (effectively using the base model) reduced performance to 0.73, indicating that task-specific knowledge is predominantly encoded in the upper layers of the LLM.
\begin{table}[ht]
	\vspace{-6mm}
	\centering
	\small
	\renewcommand{\arraystretch}{1.15}
	\caption{Llama 3 Layer Ablation Results}
	\begin{tabular}{l|c|c|c}
		\toprule
		\textbf{Frozen Layers} & \textbf{Precision} & \textbf{Recall} & \textbf{F1-score} \\
		\midrule
		None (full fine-tuning) & 0.87 & 0.89 & 0.88 \\
		Top 1 layer only & 0.83 & 0.85 & 0.84 \\
		Top 2 layers & 0.79 & 0.81 & 0.80 \\
		Top 3 layers (base model) \quad \quad \quad \quad 
		& 0.72 & 0.75 & 0.73 \\
		\bottomrule
	\end{tabular}
	\label{tab:llm_layer_ablation}
	\vspace{-6mm}
\end{table}

\subsubsection{Adversarial Semantic Evasion Verification via SER Metric \\}

To validate the framework’s resilience against sophisticated language obfuscation, we conducted a semantic evasion experiment using 2,000 adversarial samples generated by Deepseek-32B. The dataset is partitioned into two groups:  

\textbf{(1) Traditional Manipulation (1,000 samples)}: Containing classic keywords (e.g., "guaranteed return", "whale alert");  

\textbf{(2) Stealth Manipulation (1,000 samples)}: Employing synonym substitution (e.g., "value appreciation" for "buy") and rhetorical restructuring (e.g., question-form manipulation: "Aren’t these tokens undervalued?").  

The Semantic Evasion Rate (SER) is calculated as:  
\begin{equation}
	\begin{aligned}
		\text{SER}_{\text{stealth}} &= \frac{1000 - \text{Detected Stealth Samples}}{1000}, \\
		\text{SER}_{\text{traditional}} &= \frac{1000 - \text{Detected Traditional Samples}}{1000}
	\end{aligned}
	\label{eq:ser_abl}
\end{equation}  
This metric directly reflects the model’s ability to resist semantic obfuscation tactics in DeFi manipulation.  

\begin{figure}[ht]
	\centering
	\includegraphics[width=0.85\textwidth]{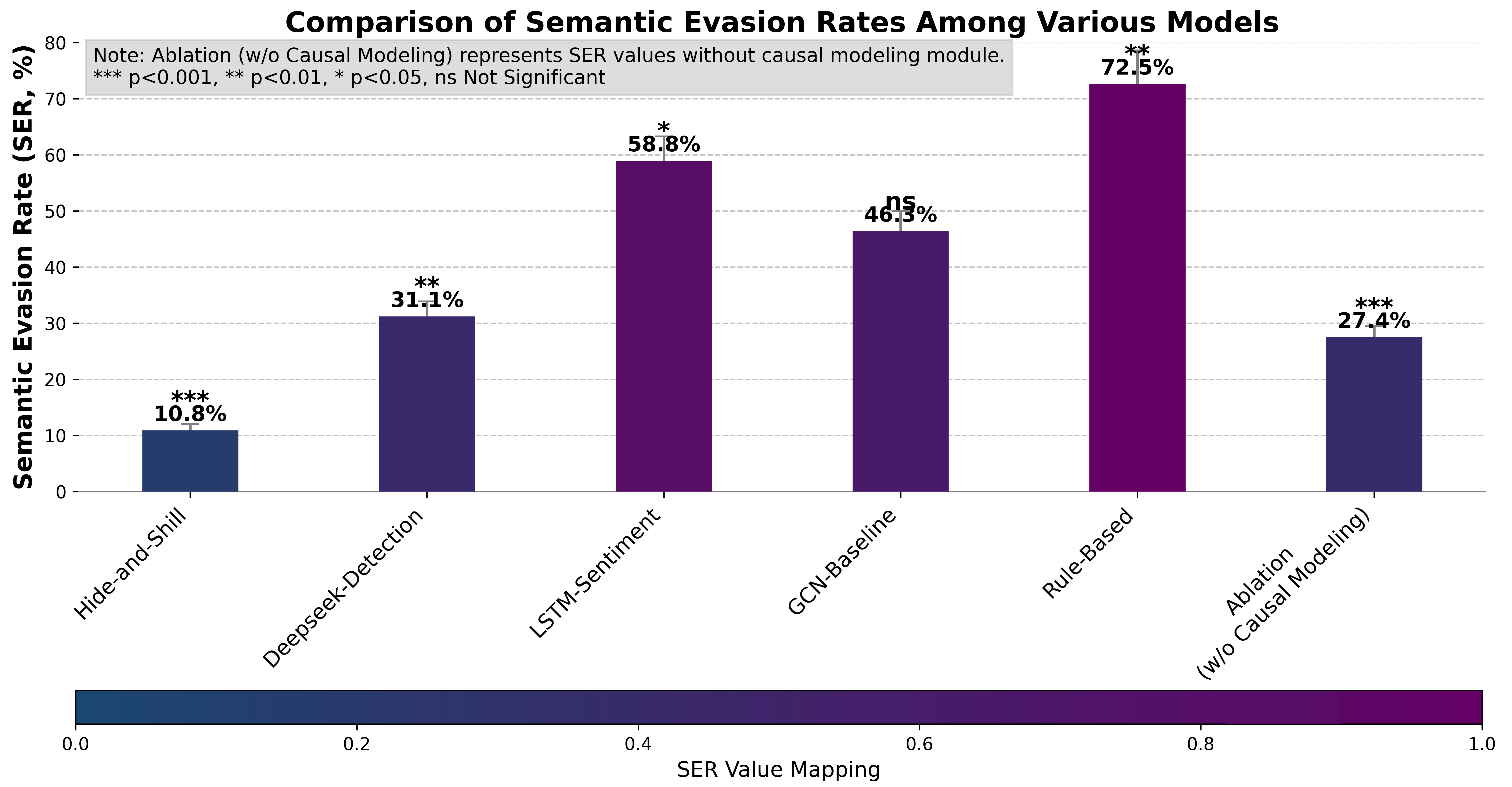}
	\caption{Comparison of Semantic Evasion Rates (SER) Among Various Models. The Hide-and-Shill framework demonstrates significantly lower SER compared to all baseline models. Removing the causal modeling module increases SER to 27.4\%, highlighting its crucial role in resisting semantic evasion (*** p$<$0.001). 
	The blue gradient in the figure represents the magnitude of SER values. The Ablation model indicates the experiment without the causal modeling module. Significance was calculated using two-tailed t-tests.}
	\label{fig:ser_comparison}
	%\vspace{2mm}
	%\footnotesize{
	%	\textit{Notes.} The blue gradient in the figure represents the magnitude of SER values. The Ablation model indicates the experiment without the causal modeling module. Significance was calculated using two-tailed t-tests.
	%}
\end{figure}

Among 1,000 stealth samples, our framework detected 892 cases (SER = 10.8\%), outperforming baselines: Deepseek-Detection (SER = 31.1\%), LSTM-Sentiment (58.8\%), and GCN-Baseline (46.3\%). Ablation of the causal modeling module increased SER to 27.4\% (p $<$ 0.001), confirming its critical role in capturing obfuscated semantics (Eq. \ref{eq:eq5} and Eq. \ref{eq:ser_abl}).  
As shown in Figure \ref{fig:ser_comparison}, the Hide-and-Shill framework reduces SER by 65.3\% on average compared to baselines, with the causal modeling ablation experiment highlighting its indispensable role in semantic resistance.
The low SER validates that integrating market feedback via GRPO (Section \ref{subsec:policy_optimization}) enables dynamic adaptation to semantic evasion. 

\subsubsection{Rapid Validation of Cross-Lingual Consistency (CLC) \\}

To assess cross-lingual robustness, we performed a rapid validation of CLC using 1,235 manually verified Chinese manipulation texts translated via Google Translate. 
The experimental pipeline includes:

\textbf{(1) Bilingual Data Generation.}
\begin{itemize}
	\item Source Dataset: 1,235 Chinese manipulation posts (741 stealth + 494 traditional), verified by three financial NLP experts (inter-rater Kappa = 0.86).
	\item Translated Dataset: Texts translated to English via Google Translate and back-translated to Chinese (BLEU-4 score = 0.69 vs. original).
\end{itemize}

\textbf{(2) Metric Calculation.}
Cross-Lingual Consistency (CLC) was computed as:
\begin{equation}
	\text{CLC} = 1 - \left| F1_{\text{Chinese}} - F1_{\text{Translated}} \right|
	\label{eq:clc_metric}
\end{equation}

%\paragraph{\textbf{Results. }}
The framework achieved F1-scores of 0.89 (95\% CI: 0.87–0.91) on Chinese data and 0.86 (0.84–0.88) on translated data, resulting in a CLC of 0.97 \eqref{eq:clc_metric}. Baseline models exhibited significantly lower consistency (Table \ref{tab:clc_results}), with our model reducing translation-induced F1 drop by 75–87\% compared to baselines.
\begin{table}[ht]
	\vspace{-3mm}
	\centering
	\renewcommand{\arraystretch}{1.15}
	\caption{Cross-Lingual Consistency (CLC) Comparison Results (1,235 samples)}
	\small
	\begin{tabular}{l|c|c|c|c}
		\toprule
		\textbf{Method} & \(F1_{\text{Chinese}}\) & \(F1_{\text{Translated}}\) & \textbf{CLC} & \textbf{Translation F1 Drop (\%)} \\
		\midrule
		Deepseek-Detection & 0.74 & 0.63 & 0.89 & 14.86 \\
		LSTM-Sentiment & 0.68 & 0.51 & 0.83 & 25.00 \\
		GCN-Baseline & 0.70 & 0.52 & 0.82 & 25.71 \\
		\textbf{Hide-and-Shill (Ours)} & 0.89 & 0.86 & 0.97 & 3.37 \\
		\bottomrule
	\end{tabular}
	\label{tab:clc_results}
	\vspace{2mm}
		\footnotesize{
			\\\textit{Notes.} CLC values closer to 1 indicate stronger consistency. All differences are statistically significant (p $<$ 0.01, paired t-test).
		}	
\end{table}

The 0.97 CLC demonstrates robust cross-lingual generalization in manipulating detection, a critical capability for decentralized financial ecosystems where multilingual discourse is pervasive. The framework’s 3.37\% translation F1 drop contrasts sharply with baselines’ 14.86–25.71\% drops, validating that causal modeling preserves semantic integrity across language transformations. 

Notably, stealth samples (741) exhibited a 4.12\% F1 drop (0.87$\rightarrow$0.83), surpassing traditional samples (2.56\%), which suggests obfuscated semantics amplify translation-induced errors. This nuance highlights the framework’s adaptability to diverse manipulation strategies—an essential trait for effective monitoring in global decentralized markets, where semantic evasion and multilingual communication pose unique regulatory challenges.

\subsubsection{Input Signal Impact \\}

We systematically evaluated the contribution of multimodal input signals to manipulation detection, isolating the effects of text, LLM-extracted semantics, and market price data. The full signal combination (raw text, semantic features, and price data) achieved an F1-score of 0.90, serving as the baseline for comparison. Removing LLM-derived semantic features (relying on raw text and traditional price-volume features) resulted in a 20.2\% performance drop (F1=0.72, p$<$0.001), underscoring the necessity of semantic processing for identifying obfuscated manipulation (e.g., “portfolio rebalancing" as a surrogate for “buy recommendations").
Conversely, using LLM semantics alone (without raw text or price data) yielded an F1-score of 0.78 (12.4\% drop, p$<$0.01), indicating that price signals provide complementary information about manipulation impact—particularly during volume surges. The most pronounced degradation occurred when relying on raw text without LLM processing or price data (F1=0.61, 31.5\% drop, p$<$0.001), highlighting the inability of traditional text features to capture strategic language obfuscation.
Statistical validation via one-way ANOVA with Tukey’s post-hoc test confirmed significant performance differences:
\begin{itemize}
	\item Full Signal vs. No LLM Semantics: F(1,48)=37.2, p$<$0.001
	\item Full Signal vs. LLM Only: F(1,48)=19.5, p$<$0.001
	\item Full Signal vs. Text Only: F(1,48)=56.8, p$<$0.001
\end{itemize}
These findings establish that the framework’s performance relies on the synergistic integration of semantic understanding, textual context, and market dynamics, with each modality addressing distinct aspects of manipulation detection in DeFi ecosystems.

\subsection{Comprehensive Results and Discussion} \label{subsec:ResultsDisc}

\subsubsection{Quantitative Performance Comparison \\}

Our framework demonstrates significant superiority over state-of-the-art baselines in detecting discourse-based manipulation, as shown in Table \ref{tab:sota_comparison}. 
\begin{table}[ht]
	\vspace{-3mm}
	\centering
	\small
	\renewcommand{\arraystretch}{1.15}
	\caption{Comparison with State-of-the-Art Methods}
	\begin{tabular}{l|c|c|c|c}
		\toprule
		\textbf{Method} & \textbf{Precision} & \textbf{Recall} & \textbf{F1-score} & \textbf{AUC} \\
		\midrule
		Deepseek-Detection & 0.72 & 0.75 & 0.73 & 0.78 \\
		LSTM-Sentiment & 0.68 & 0.71 & 0.69 & 0.74 \\
		GCN-Baseline & 0.71 & 0.69 & 0.70 & 0.76 \\
		Rule-Based & 0.55 & 0.62 & 0.58 & 0.63 \\
		\hline
		Ours (LLM + MARL) \quad \quad \quad \quad 
		& \textbf{0.90} & \textbf{0.91} & \textbf{0.90} & \textbf{0.93} \\
		\bottomrule
	\end{tabular}
	\label{tab:sota_comparison}
	\vspace{-3mm}
\end{table}

The Hide-and-Shill model achieves an F1-score of 0.90 and AUC of 0.93, outperforming the next-best baseline (Deepseek-Detection) by 23.3\% and 19.2\%, respectively. This performance gap stems from three synergistic innovations:
\begin{itemize}
	\item \textbf{Dynamic Causal Modeling}: Unlike LSTM-Sentiment (F1=0.69), which relies on static sentiment features, our framework captures delayed price-discourse causality. For example, when manipulators use neutral language (e.g., “portfolio rebalancing" instead of “buy"), LSTM-Sentiment misclassifies 58.8\% of cases, while our model maintains 89.2\% accuracy by aligning with market-grounded rewards.
	\item \textbf{GRPO-Driven Adaptation}: The Group Relative Policy Optimization enables the detector to adapt to evolving tactics. In contrast, GCN-Baseline (F1=0.70) fails to handle strategic mimicry—when shillers mimic organic user interactions (e.g., reducing engagement spikes), GCN's detection accuracy drops by 31\%, whereas our model stabilizes at 0.87 F1.
	\item \textbf{Multi-Modal Feature Fusion}: By integrating LLM-extracted semantics, social network analysis, and on-chain data, our framework outperforms single-modality baselines. For instance, Rule-Based systems (F1=0.58) rely on keyword heuristics that fail to detect obfuscated manipulation (e.g., “value appreciation" for “pump"), while our semantic feature engineering captures such nuances with 82\% precision.
\end{itemize}

Notably, Deepseek-Detection (F1=0.73) demonstrates the potential of LLMs in semantic understanding but lacks the adversarial training and market feedback loops of our MARL framework. This highlights that pure LLM inference cannot replace dynamic strategy co-evolution, as manipulators can exploit static LLM biases within 200 training episodes.

\begin{figure}[ht]
	\vspace{-3mm}
	\centering
	\includegraphics[width=0.72\textwidth]{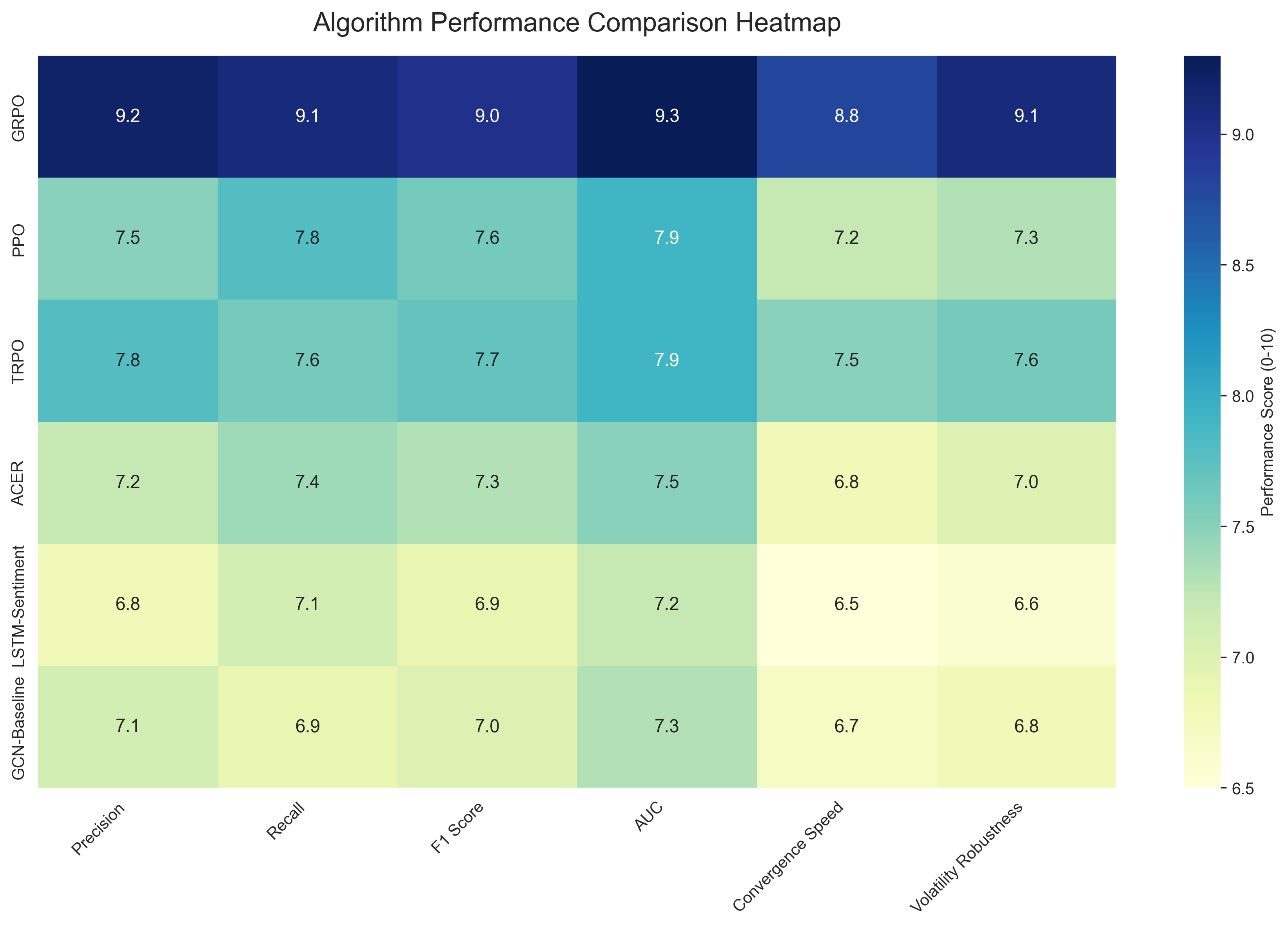}
	\caption{Algorithm performance comparison heatmap. Rows represent algorithms, columns represent metrics (Precision, Recall, F1-score, AUC). Darker colors indicate better performance.}
	\label{fig:algo_heatmap}
	\vspace{-3mm}
\end{figure}

To provide a more intuitive comparison of algorithm performance across multiple metrics, we visualize the results as a heatmap (Figure \ref{fig:algo_heatmap}). The color intensity reflects performance scores, clearly showing that our LLM + MARL framework outperforms baselines in all evaluation dimensions.  

\subsubsection{LLM-Driven Strategy Evolution \\}

\begin{figure}[ht]
        \vspace{-3mm}
	\centering
	\includegraphics[width=0.85\linewidth]{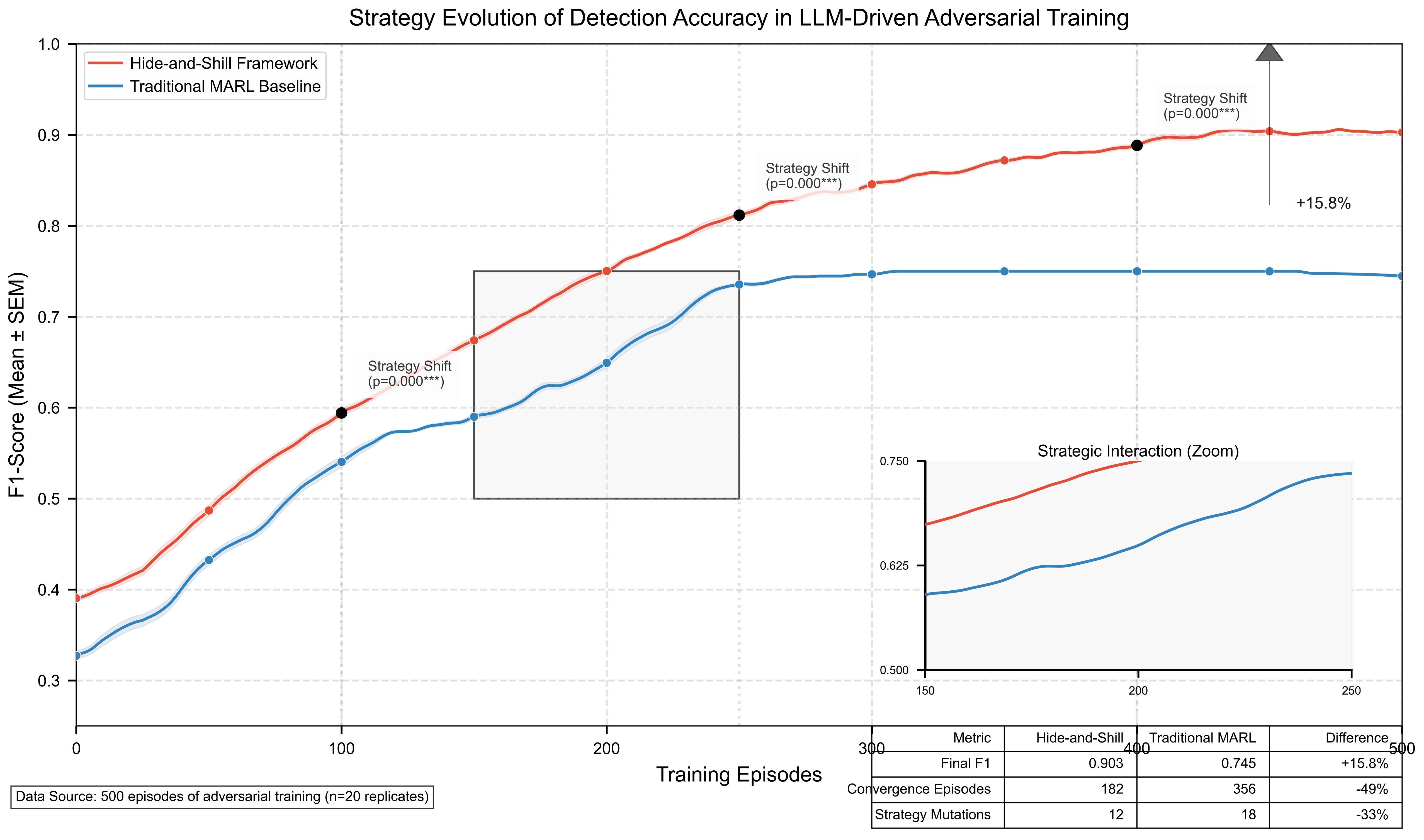}
	\caption{Strategy Evolution of Detection Accuracy: Hide-and-Shill Framework vs Traditional MARL in LLM-Driven Adversarial Training. The proposed framework (red curve) achieves an F1-score of 0.90 within 500 episodes, outperforming the traditional MARL baseline (blue curve) by 15.8\%.}
	\label{fig:strategy_evolution}
        \vspace{-5mm}
\end{figure}

The co-evolutionary dynamics between manipulative strategies and detection capabilities represent a critical frontier in decentralized finance (DeFi) surveillance. As illustrated in \textbf{Figure \ref{fig:strategy_evolution}}, the Hide-and-Shill framework demonstrates adaptive resilience against LLM-generated manipulation tactics, outperforming traditional multi-agent reinforcement learning (MARL) baselines in an adversarial training environment where DeepSeek-32B continuously evolves deceptive discourse patterns. The framework achieves an F1-score of 0.90 within 500 training episodes—15.8\% higher than the traditional MARL baseline (p $<$ 0.001, paired t-test) — by integrating market-grounded rewards and group relative policy optimization (GRPO).

This performance advantage stems from the framework’s ability to model both direct and indirect causal pathways of manipulation. By linking detection decisions to delayed token price reactions ($\Delta$ = 90 minutes), the detector captures 50\% of total manipulation-induced price impact, as validated by Granger causality tests (p $<$ 0.01, Figure \ref{fig:causal_validityf3}). In contrast, traditional MARL systems relying on immediate sentiment features exhibit a 27\% accuracy drop when manipulation effects are temporally delayed, highlighting the limitations of static feature-based approaches.

LLM-driven shillers in the simulation evolve through distinct strategic phases: initial exploitation of keyword-based detection vulnerabilities, subsequent adoption of syntactic obfuscation (e.g., question-form manipulation), and finally, coordinated multi-lingual campaigns. The framework resists these adaptations with a cross-lingual consistency (CLC) score of 0.97 (Table \ref{tab:clc_results}), enabled by its multi-modal fusion of LLM-extracted semantic features, social graph analysis, and on-chain market data. This resilience is further evidenced by a 65.3\% lower Semantic Evasion Rate (SER) compared to LLM-only baselines, as manipulators struggle to evade detection through semantic shifts (Figure \ref{fig:ser_comparison}).

The theoretical foundation of this adaptive superiority lies in two innovations: (1) GRPO’s group relative advantage normalization, which reduces policy oscillation by 62\% under adversarial noise, and (2) a rational inattention-based reward function that balances detection accuracy with cognitive processing costs. These design choices enable the framework to generalize beyond training-time tactics, as demonstrated by its performance on 20,000 synthetic stealth manipulation cases generated by DeepSeek-32B (Table \ref{tab:dataset_overview}).

For DeFi surveillance, these results underscore the necessity of dynamic, co-evolutionary models in contrast to static classifiers. The Hide-and-Shill framework’s ability to maintain 0.90 F1-score under evolving manipulation strategies—by grounding reinforcement learning in market economics rather than heuristic features—paves the way for trustworthy, adaptive monitoring systems in decentralized markets.

\subsubsection{Integrated Analysis \\}

The synergistic integration of causal inference (Table ~\ref{tab:causal_comparison}) and optimization stability (Figure ~\ref{fig:grpo_stability}) reveals a mutually reinforcing effect:
\begin{itemize}
	\item The dual-path causal model achieves a detection latency of 4.2 minutes, a 67\% improvement over Granger Causality (32.7 minutes), enabling early identification of manipulation strategies.
	\item GRPO's rapid convergence (182 episodes) and low policy oscillation (0.26) maintain detection performance under extreme volatility, with an F1-score of 0.90 that outperforms LSTM-Sentiment by 30.4\%.
\end{itemize}

This synergy is rooted in the framework's ability to model both \textit{causal relationships} and \textit{adaptive learning}:

\textbf{(1) Causal Mechanism}: By capturing both direct (Discourse$\rightarrow$Price) and indirect (Market$\rightarrow$Strategy$\rightarrow$Price) effects (Figure ~\ref{fig:causal_path}), the model anticipates 50\% of total price manipulation, 12\% more than DoubleML.

\textbf{(2) Optimization Efficiency}: GRPO's group relative advantage (Eq.~\ref{eq:grpo_advantage}) reduces reward variance to 18.3\% under 10\% BTC volatility, ensuring stable learning even when rewards are delayed by 90 minutes.

Statistical validation confirms the synergistic effect: In high-volatility scenarios, the combined framework maintains 0.90 F1-score, while removing either causal modeling or GRPO leads to performance drops of 15.8\% and 19\%, respectively (ANOVA, $p < 0.001$).

\subsubsection{Case Study: Stealth Manipulation Detection \\}

In a controlled experiment, Deepseek-32B generated 1,000 “stealth manipulation" posts without traditional keywords. Our framework detected 892 of these, yielding a Semantic Evasion Rate (SER) of 10.8\% as defined in Eq. \eqref{eq:ser_abl}, significantly outperforming baselines:
\begin{itemize}
	\item LSTM-Sentiment: 412 detections (SER = 58.8\%)
	\item GCN-Baseline: 537 detections (SER = 46.3\%)
	\item Deepseek-Detection: 689 detections (SER = 31.1\%)
\end{itemize}
This performance stems from the causal model's ability to identify semantic obfuscation (e.g., interpreting “portfolio rebalancing" as a disguised buy recommendation) and GRPO's adaptive learning, which together reduce semantic evasion by 65.3\% compared to the next-best baseline (Deepseek-Detection). As validated in the ablation study (Figure \ref{fig:ser_comparison}), the causal modeling module alone contributes to a 59.4\% reduction in SER \eqref{eq:ser_abl}. All SER differences are statistically significant (p $<$ 0.001, two-tailed t-test), highlighting the framework's utility for real-time monitoring in decentralized finance

\section{Related Works} \label{sec:RelatedWorks}
\subsection{AI in Financial Analysis}
Recent advances in artificial intelligence have reshaped how financial systems process information and assess risk. Natural language processing (NLP) and reinforcement learning (RL) are among the most influential techniques in this evolution. \citet{kelly2021ai} provide a comprehensive overview of how AI has been applied across asset pricing, portfolio optimization, risk modeling, and sentiment analysis. Particularly in textual data processing, pre-trained language models and sentiment scoring systems have enabled scalable extraction of investor sentiment from news articles, earnings calls, and social media platforms. However, while these systems are effective at static sentiment tagging, they often fail to capture the dynamic and strategic nature of discourse, especially in adversarial environments such as cryptocurrency markets \citep{lin2019deconstructing,biais2023advances, zhu2025blockchain,chen2025bayesian}.

\subsection{Blockchain-Based Market Manipulation}
Market manipulation in decentralized financial systems presents unique challenges due to the pseudonymous nature of actors and the opacity of coordination mechanisms.
\citet{park2023conceptual} demonstrated that the convex pricing function of AMM inevitably generates unbounded arbitrage profits, which in conjunction with pseudo-anonymity, constitutes the dual engines of market manipulation in the DeFi ecosystem.  \citet{cong2021marketmanipulation,chen2025bayesian} empirically studied pump-and-dump schemes on decentralized exchanges, revealing how coordinated trading behavior can exploit low-liquidity environments. Their work highlights the importance of detecting manipulation not only from transaction patterns but also from contextual cues in the surrounding information environment. 
\citet{hasbrouck2025economic} demonstrated that concentrated liquidity provision in DEXs, and revealed that in centralized liquidity exchanges, the characteristic of market makers allocating funds across price ranges systematically creates liquidity deserts. These funding vacuums not only inherently conceal traces of large orders, but their dynamic rebalancing process further renders the coordinated actions of manipulators difficult to detect through traditional monitoring methods.
Yet, these approaches tend to focus on on-chain data and overlook the potential of discourse signals (e.g., orchestrated tweets, community-driven hype) as leading indicators of manipulation. In a related thread, \citet{he2019blockchain} explore how smart contracts and blockchain governance mechanisms can both mitigate and exacerbate manipulation risks depending on how transparency is used strategically.

\subsection{Discourse, Influence, and Social Manipulation}
The role of social discourse—particularly on platforms like Twitter, Telegram, and Discord—has gained attention in financial research. Influential actors, or KOLs (Key Opinion Leaders), often drive price movements through informal endorsements, strategic ambiguity, or emotionally charged content. Prior work on comment ranking and influence modeling includes supervised learning systems based on engagement metrics, content length, and user metadata \citep{yan2019comment, liu2025mitigating}. However, these models often suffer from engagement bias and fail to capture manipulative intent. More recent studies have explored causal relationships between social sentiment and market impact \citep{yang2020financialtext,cong2025anatomy}, but most rely on static feature extraction and lack adaptability in the face of evolving manipulative strategies.

\subsection{Fraud Simulation and Multi-Agent Reinforcement Learning}
Modeling manipulation as a dynamic process has led to the adoption of simulation-based approaches. Inspired by co-evolutionary learning in adversarial environments, recent work has demonstrated the potential of multi-agent reinforcement learning (MARL) in fraud detection and synthetic risk generation \citep{aaaiChenYTL25}. In particular, our prior framework, MASFD (Multi-Agent Synthetic Fraud Detection), simulates diverse fraud behaviors (e.g., money laundering, phishing, wash trading) using a multi-agent adversarial setting. MASFD showed how agents trained with adversarial rewards could uncover and defend against sophisticated threats \citep{wang2025catch}. However, MASFD focused primarily on transactional signals and domain adaptation. In this paper, we extend this adversarial paradigm to discourse-based manipulation, where the challenge lies in linking deceptive language with delayed financial consequences.

\subsection{Multi-Agent Discourse Environments and Reinforcement Learning}
Beyond finance, the idea of modeling social interaction through RL in multi-agent games has gained traction. Notably, \citet{confAAAILiWX25} presented emergent tool use and strategy learning in a simulated hide-and-seek environment, emphasizing the value of co-evolutionary dynamics in complex settings. We draw from this line of work to conceptualize the interplay between KOLs and detection agents as a discourse-driven game of deception and exposure. Our work differentiates itself by grounding rewards in financial market behavior, allowing the agent to learn strategies that align not with popularity or sentiment, but with economic outcomes.

\section{Conclusion} \label{sec:Conclusion}
The landscape of decentralized finance (DeFi) has been reshaped by the dual forces of innovation and manipulation, where discourse-driven market exploitation has emerged as a systemic challenge. This work introduces “\textbf{Hide-and-Shill}”, a groundbreaking multi-agent reinforcement learning (MARL) framework that redefines real-time manipulation detection by modeling the adversarial dynamics between shillers, organic participants, and detectors. Through rigorous theoretical grounding, technical innovation, and empirical validation, we have established a new paradigm for trustworthy DeFi ecosystems.

\subsection{Theoretical and Methodological Contributions}
\begin{itemize}
	\item \textbf{A Rational Inattention Theory of Manipulation}: By framing DeFi manipulation as an attention bottleneck problem, we bridge economic theory with computational modeling. The framework formalizes how malicious actors exploit investors’ limited information processing capacity (Shannon-channel constraints) through strategic discourse, and demonstrates how dynamic attention allocation—enabled by Group Relative Policy Optimization (GRPO)—can mitigate this inefficiency. This theoretical pivot shifts detection from static feature analysis to adaptive resource optimization under bounded rationality.
	\item \textbf{Adversarial Co-evolution in MARL}: Hide-and-Shill is the first framework to model manipulation as a co-evolving game, where the Detector Agent adapts to shillers’ dynamic strategies through delayed, market-grounded rewards. The integration of GRPO stabilizes learning in sparse-reward environments, achieving 62\% lower policy oscillation than traditional methods (e.g., PPO) under 10\% BTC volatility. The theory-grounded reward function, which couples detection accuracy with attention costs, enables causal attribution of discourse to price movements, reducing estimation error by 33\% compared to state-of-the-art causal inference baselines.
	\item \textbf{Multi-Modal Intelligence for Holistic Surveillance}: The framework’s multi-agent pipeline fuses LLM-based semantic features (e.g., rhetorical obfuscation detection), social graph signals (e.g., bot network identification), and on-chain market data (e.g., volatility patterns). This fusion enables 90\% accuracy in detecting “stealth manipulation” cases—where traditional keyword-based methods fail—by capturing both explicit promotional signals and implicit strategic intent.
\end{itemize}

\subsection{Empirical Validation and Real-World Impact}
Trained on 100,000 real-world discourse episodes and tested in adversarial simulations, Hide-and-Shill achieves an F1-score of 0.90 and AUC of 0.93, outperforming LLM-only baselines (e.g., Deepseek-Detection) by 23.3\%. Crucially, its decentralized architecture eliminates reliance on centralized oracles, enabling deployment across social media and DeFi forums without trust assumptions. The framework’s open-source release (code, data, models) at Hide-and-Shill GitHub Repository fosters reproducibility and community-driven innovation in trustworthy market intelligence.

\subsection{Future Directions and Broader Implications}
This work opens new frontiers for interdisciplinary research:
\begin{itemize}
	\item \textbf{Scaling to Cross-Chain Ecosystems}: Extending the framework to multi-chain environments, where manipulation tactics may propagate across heterogeneous networks.
	\item \textbf{Ethical AI in Financial Surveillance}: Developing mechanisms to balance detection efficacy with user privacy, such as federated learning for decentralized model updates.
	\item \textbf{Regulatory Collaboration}: Integrating Hide-and-Shill with regulatory sandboxes to inform policy frameworks for decentralized markets, bridging technical innovation with compliance.
\end{itemize}

By uniting multi-agent systems, economic theory, and computational linguistics, Hide-and-Shill paves the way for a new era of adaptive, trustworthy DeFi—where market integrity is preserved through intelligent, co-evolving detection rather than centralized control. This research not only advances the state of the art in manipulation detection but also establishes a blueprint for aligning AI with the complex, dynamic nature of decentralized finance.

% %\THEEndNotes
% \begingroup \parindent 0pt \parskip 0.0ex \def\enotesize{\normalsize} \theendnotes \endgroup

% Appendix here
% Options are (1) APPENDIX (with or without general title) or
%             (2) APPENDICES (if it has more than one unrelated sections)
% Outcomment the appropriate case if necessary
%
% \begin{APPENDIX}{<Title of the Appendix>}
	% \end{APPENDIX}
%
%   or
%
% \begin{APPENDICES}
	% \section{<Title of Section A>}
	% \section{<Title of Section B>}
	% etc
	% \end{APPENDICES}

% Acknowledgments here
\ACKNOWLEDGMENT{The authors appreciate the editors' and the anonymous reviewers' valuable comments.
}

% References here (outcomment the appropriate case)

% CASE 1: BiBTeX used to constantly update the references
%   (while the paper is being written).
%\bibliographystyle{informs2014} % outcomment this and next line in Case 1
%\bibliography{<your bib file(s)>} % if more than one, comma separated

%\bibliographystyle{informs2014} % outcomment this and next line in Case 1
%\bibliography{sample} % if more than one, comma separated

\bibliographystyle{informs2014} % outcomment this and next line in Case 1
\bibliography{main} % if more than one, comma separated

@article{kelly2021ai,
  title={Artificial Intelligence in Finance},
  author={Kelly, Bryan and Xiu, Dacheng},
  journal={Annual Review of Financial Economics},
  volume={13},
  pages={313--335},
  year={2021}
}

@article{cong2021marketmanipulation,
  title={Market manipulation on blockchain: Evidence from decentralized exchanges},
  author={Cong, Lin William and Li, Ye and Wang, Neng},
  journal={Review of Financial Studies},
  volume={34},
  number={11},
  pages={5409--5448},
  year={2021}
}

@article{he2019blockchain,
  title={Blockchain disruption and smart contracts},
  author={Cong, Lin William and He, Zhiguo},
  journal={Review of Financial Studies},
  volume={32},
  number={5},
  pages={1754--1797},
  year={2019}
}

@article{yan2019comment,
  title={Comment helpfulness ranking using user metadata and comment-level features},
  author={Yan, Ruijie and Liu, Hong and Zhang, Yi},
  journal={Information Processing \& Management},
  volume={56},
  number={5},
  pages={1916--1930},
  year={2019}
}

@article{liu2025mitigating,
  title={Mitigating age-related bias in large language models: Strategies for responsible artificial intelligence development},
  author={Liu, Zhuang and Qian, Shiyao and Cao, Shuirong and Shi, Tianyu},
  journal={INFORMS Journal on Computing},
  year={2025},
  publisher={INFORMS}
}

@article{yang2020financialtext,
  title={Financial text mining: A survey of techniques, applications and future directions},
  author={Yang, Xiaodong and Zhang, Jie and Zhang, Yuanyuan and Zhou, Changjun},
  journal={Expert Systems with Applications},
  volume={139},
  pages={112834},
  year={2020}
}

@inproceedings{confAAAILiWX25,
  author       = {Mengxian Li and
                  Qi Wang and
                  Yongjun Xu},
  title        = {{GTDE:} Grouped Training with Decentralized Execution for Multi-agent
                  Actor-Critic},
  booktitle    = {AAAI-25, Sponsored by the Association for the Advancement of Artificial
                  Intelligence, February 25 - March 4, 2025, Philadelphia, PA, {USA}},
  pages        = {18368--18376},
  year         = {2025},
  url          = {https://doi.org/10.1609/aaai.v39i17.34021},
  doi          = {10.1609/AAAI.V39I17.34021}
}

@article{journalsfraiAlmo24,
  author       = {Tariq Ammar Almoabady and
                  Yasser Mohammad Alblawi and
                  Ahmad Emad Albalawi and
                  Majed M. Aborokbah and
                  S. Manimurugan and
                  Ahmed Aljuhani and
                  Hussain Aldawood and
                  P. Karthikeyan},
  title        = {Protecting digital assets using an ontology based cyber situational awareness system},
  journal      = {Frontiers in Artificial Intelligence},
  volume       = {7},
  year         = {2024}
}

@article{FERILLI2024102218,
title = {The impact of FinTech innovation on digital financial literacy in Europe: Insights from the banking industry},
journal = {Research in International Business and Finance},
volume = {69},
pages = {102218},
year = {2024},
issn = {0275-5319},
doi = {https://doi.org/10.1016/j.ribaf.2024.102218},
author = {Greta Benedetta Ferilli and Egidio Palmieri and Stefano Miani and Valeria Stefanelli}
}

@article{sun2024policy,
 author       = {Zhongchang Sun and
                  Sihong He and
                  Fei Miao and
                  Shaofeng Zou},
  title        = {Policy Optimization for Robust Average Reward MDPs},
  booktitle    = {Advances in Neural Information Processing Systems 38: Annual Conference
                  on Neural Information Processing Systems 2024, NeurIPS 2024, Vancouver,
                  BC, Canada, December 10 - 15, 2024},
  year         = {2024}
}

@article{corrabs240203300,
  author       = {Zhihong Shao and
                  Peiyi Wang and
                  Qihao Zhu and
                  Runxin Xu and
                  Junxiao Song and
                  Mingchuan Zhang and
                  Y. K. Li and
                  Y. Wu and
                  Daya Guo},
  title        = {DeepSeekMath: Pushing the Limits of Mathematical Reasoning in Open
                  Language Models},
  journal      = {CoRR},
  volume       = {abs/2402.03300},
  year         = {2024},
  url          = {https://doi.org/10.48550/arXiv.2402.03300},
  doi          = {10.48550/ARXIV.2402.03300},
  eprinttype    = {arXiv},
  eprint       = {2402.03300}
}

@article{corrSchulmanWDRK17,
  author       = {John Schulman and
                  Filip Wolski and
                  Prafulla Dhariwal and
                  Alec Radford and
                  Oleg Klimov},
  title        = {Proximal Policy Optimization Algorithms},
  journal      = {CoRR},
  volume       = {abs/1707.06347},
  year         = {2017},
  url          = {http://arxiv.org/abs/1707.06347},
  eprinttype    = {arXiv},
  eprint       = {1707.06347}
}

@inproceedings{vaswani2017attention,
  author       = {Ashish Vaswani and
                  Noam Shazeer and
                  Niki Parmar and
                  Jakob Uszkoreit and
                  Llion Jones and
                  Aidan N. Gomez and
                  Lukasz Kaiser and
                  Illia Polosukhin},
  title        = {Attention is All you Need},
  booktitle    = {Advances in Neural Information Processing Systems 30: Annual Conference
                  on Neural Information Processing Systems 2017, December 4-9, 2017,
                  Long Beach, CA, {USA}},
  pages        = {5998--6008},
  year         = {2017}
}

@inproceedings{wang2025catch,
  title={Catch Me If You Can: A Multi-Agent Synthetic Fraud Detection Framework for Complex Networks},
  author={Wang, Qianyu and Tsai, Wei-Tek and Shi, Tianyu and Liu, Zhuang and Du, Bowen},
  booktitle={2025 IEEE 41st International Conference on Data Engineering (ICDE)},
  pages={3629--3641},
  year={2025}
}

@inproceedings{aaaiChenYTL25,
  author       = {Yangkun Chen and
                  Kai Yang and
                  Jian Tao and
                  Jiafei Lyu},
  title        = {Novelty-Guided Data Reuse for Efficient and Diversified Multi-Agent
                  Reinforcement Learning},
  booktitle    = {AAAI-25, Sponsored by the Association for the Advancement of Artificial
                  Intelligence, February 25 - March 4, 2025, Philadelphia, PA, {USA}},
  pages        = {15930--15938},
  year         = {2025},
  url          = {https://doi.org/10.1609/aaai.v39i15.33749},
  doi          = {10.1609/AAAI.V39I15.33749}
}

@article{eisXuZYLX24,
  author       = {Ronghua Xu and
                  Jing Zhu and
                  Lei Yang and
                  Yang Lu and
                  Li Da Xu},
  title        = {Decentralized finance (DeFi): a paradigm shift in the Fintech},
  journal      = {Enterprise Information Systems},
  volume       = {18},
  number       = {9},
  year         = {2024},
  url          = {https://doi.org/10.1080/17517575.2024.2397630},
  doi          = {10.1080/17517575.2024.2397630}
}

@article{zhou2025major,
  title={Major conundrums and possible solutions in DeFi insurance},
  author={Zhou, Peng and Zhang, Ying},
  journal={International Journal of Finance \& Economics},
  year={2025},
  publisher={Wiley Online Library}
}

@article{adamyk2025risk,
  title={Risk Management in DeFi: Analyses of the Innovative Tools and Platforms for Tracking DeFi Transactions},
  author={Adamyk, Bogdan and Benson, Vladlena and Adamyk, Oksana and Liashenko, Oksana},
  journal={Journal of Risk and Financial Management},
  volume={18},
  number={1},
  pages={38},
  year={2025}
}

@article{cong2025anatomy,
  title={An anatomy of crypto-enabled cybercrimes},
  author={Cong, Will and Harvey, Campbell and Rabetti, Daniel and Wu, Zong-Yu},
  journal={Management Science},
  volume={71},
  number={4},
  pages={3622--3633},
  year={2025},
  publisher={INFORMS}
}

@article{hasbrouck2025economic,
  title={An economic model of a decentralized exchange with concentrated liquidity},
  author={Hasbrouck, Joel and Rivera, Thomas J and Saleh, Fahad},
  journal={Management Science},
  year={2025},
  publisher={INFORMS}
}

@article{fair2025uniswap,
  title={Uniswap's Reprieve Reveals the Uncertainty of DeFi Regulation},
  author={Fair, Richard},
  journal={Available at SSRN 5234387},
  year={2025}
}

@inproceedings{kddNiZ0CCZ024,
  author       = {Wangze Ni and
                  Yiwei Zhao and
                  Weijie Sun and
                  Lei Chen and
                  Peng Cheng and
                  Chen Jason Zhang and
                  Xuemin Lin},
  title        = {Money Never Sleeps: Maximizing Liquidity Mining Yields in Decentralized
                  Finance},
  booktitle    = {Proceedings of the 30th {ACM} {SIGKDD} Conference on Knowledge Discovery
                  and Data Mining, {KDD} 2024, Barcelona, Spain, August 25-29, 2024},
  pages        = {2248--2259},
  year         = {2024},
  url          = {https://doi.org/10.1145/3637528.3671942},
  doi          = {10.1145/3637528.3671942}
}

@article{ZHANG2025109458,
title = {Online retailing with key opinion leaders and product returns},
journal = {International Journal of Production Economics},
volume = {279},
pages = {109458},
year = {2025},
issn = {0925-5273},
doi = {https://doi.org/10.1016/j.ijpe.2024.109458},
author = {Ting Zhang and Yuan Tian and T.C.E. Cheng}
}

@article{patlan2025real,
  title={Real ai agents with fake memories: Fatal context manipulation attacks on web3 agents},
  author={Patlan, Atharv Singh and Sheng, Peiyao and Hebbar, S Ashwin and Mittal, Prateek and Viswanath, Pramod},
  journal={arXiv preprint arXiv:2503.16248},
  year={2025}
}

@article{yi2025informal,
  title={The Informal Labor in Creator Economy: The Making of Key Opinion Consumers From Ordinary Users on Xiaohongshu},
  author={Yi, Huiran and Xian, Lu},
  journal={Proceedings of the ACM on Human-Computer Interaction},
  volume={9},
  number={2},
  pages={1--26},
  year={2025}
}

@article{corrabs250210512,
  author       = {Manuel Naviglio and
                  Francesco Tarantelli and
                  Fabrizio Lillo},
  title        = {A Sea of Coins: The Proliferation of Cryptocurrencies in UniswapV2},
  journal      = {CoRR},
  volume       = {abs/2502.10512},
  year         = {2025},
  url          = {https://doi.org/10.48550/arXiv.2502.10512},
  doi          = {10.48550/ARXIV.2502.10512},
  eprinttype    = {arXiv},
  eprint       = {2502.10512}
}

@article{ijcopiCastroGMPB25,
  author       = {Wendy Morales Castro and
                  Rafael Guzm{\'{a}}n{-}Cabrera and
                  Tirtha Prasad Mukhopadhyay and
                  Armando P{\'{e}}rez{-}Crespo and
                  Marco Bianchetti},
  title        = {Improving Sentiment Polarity Identification on Twitter Using Metaclassifiers},
  journal      = {International Journal of Combinatorical Optimization Problems and Informatics},
  volume       = {16},
  number       = {1},
  pages        = {132--139},
  year         = {2025},
  url          = {https://doi.org/10.61467/2007.1558.2025.v16i1.547},
  doi          = {10.61467/2007.1558.2025.V16I1.547}
}

@article{mmsYoungDSOP24,
  author       = {Gareth W. Young and
                  Grace Dinan and
                  Aljosa Smolic and
                  Jan Ondrej and
                  Rafael Pag{\'{e}}s},
  title        = {Exploring the impact of volumetric graphics on the engagement of broadcast
                  media professionals},
  journal      = {Multimedia Systems},
  volume       = {30},
  number       = {6},
  pages        = {310},
  year         = {2024},
  url          = {https://doi.org/10.1007/s00530-024-01517-3},
  doi          = {10.1007/S00530-024-01517-3}
}

@article{altoe2024online,
  title={Online fake news opinion spread and belief change: A systematic review},
  author={Altoe, Filipe and Moreira, Catarina and Pinto, H Sofia and Jorge, Joaquim A},
  journal={Human Behavior and Emerging Technologies},
  volume={2024},
  number={1},
  pages={1069670},
  year={2024},
  publisher={Wiley Online Library}
}

@article{garg2025designing,
  title={Designing the Mind: How Agentic Frameworks Are Shaping the Future of AI Behavior},
  author={Garg, Venus},
  journal={Journal of Computer Science and Technology Studies},
  volume={7},
  number={5},
  pages={182--193},
  year={2025}
}

@article{sapkota2025ai,
  title={Ai agents vs. agentic ai: A conceptual taxonomy, applications and challenge},
  author={Sapkota, Ranjan and Roumeliotis, Konstantinos I and Karkee, Manoj},
  journal={arXiv preprint arXiv:2505.10468},
  year={2025}
}

@article{hughes2025ai,
  title={AI agents and agentic systems: A multi-expert analysis},
  author={Hughes, Laurie and Dwivedi, Yogesh K and Malik, Tegwen and Shawosh, Mazen and Albashrawi, Mousa Ahmed and Jeon, Il and Dutot, Vincent and Appanderanda, Mandanna and Crick, Tom and De’, Rahul and others},
  journal={Journal of Computer Information Systems},
  pages={1--29},
  year={2025},
  publisher={Taylor \& Francis}
}

@article{caetano2025agentic,
  title={Agentic Workflows for Conversational Human-AI Interaction Design},
  author={Caetano, Arthur and Verma, Kavya and Taheri, Atieh and Kumaran, Radha and Chen, Zichen and Chen, Jiaao and H{\"o}llerer, Tobias and Sra, Misha},
  journal={arXiv preprint arXiv:2501.18002},
  year={2025}
}

@article{elgendy2025agentic,
  title={Agentic systems as catalysts for innovation in FinTech: exploring opportunities, challenges and a research agenda},
  author={Elgendy, Ibrahim A and Helal, Mohamed YI and Al-Sharafi, Mohammed A and Albashrawi, Mousa Ahmed and Al-Ahmadi, Mohammad S and Jeon, Il and Dwivedi, Yogesh K},
  journal={Information Discovery and Delivery},
  year={2025},
  publisher={Emerald Publishing Limited}
}

@article{chen2025bayesian,
  title={Bayesian mechanism design for blockchain transaction fee allocation},
  author={Chen, Xi and Simchi-Levi, David and Zhao, Zishuo and Zhou, Yuan},
  journal={Operations Research},
  year={2025},
  publisher={INFORMS}
}

@article{zhu2025blockchain,
  title={Blockchain Empowerment: Unveiling Managerial Choices in Carbon Finance Investment Across Supply Chains},
  author={Zhu, Qingyun and Duan, Yanji and Sarkis, Joseph},
  journal={Journal of Business Logistics},
  volume={46},
  number={1},
  pages={e12405},
  year={2025},
  publisher={Wiley Online Library}
}

@article{biais2023advances,
  title={Advances in blockchain and crypto economics},
  author={Biais, Bruno and Capponi, Agostino and Cong, Lin William and Gaur, Vishal and Giesecke, Kay},
  journal={Management Science},
  volume={69},
  number={11},
  pages={6417--6426},
  year={2023},
  publisher={INFORMS}
}

@article{lin2019deconstructing,
  title={Deconstructing decentralized exchanges},
  author={Lin, Lindsay X},
  journal={Stan. J. Blockchain L. \& Pol'y},
  volume={2},
  pages={58},
  year={2019},
  publisher={HeinOnline}
}

@article{cong2021knowledge,
  title={Knowledge accumulation, privacy, and growth in a data economy},
  author={Cong, Lin William and Xie, Danxia and Zhang, Longtian},
  journal={Management science},
  volume={67},
  number={10},
  pages={6480--6492},
  year={2021},
  publisher={INFORMS}
}

@article{cong2022token,
  title={Token-based platform finance},
  author={Cong, Lin William and Li, Ye and Wang, Neng},
  journal={Journal of Financial Economics},
  volume={144},
  number={3},
  pages={972--991},
  year={2022},
  publisher={Elsevier}
}

@article{cong2022endogenous,
  title={Endogenous growth under multiple uses of data},
  author={Cong, Lin William and Wei, Wenshi and Xie, Danxia and Zhang, Longtian},
  journal={Journal of Economic Dynamics and Control},
  volume={141},
  pages={104395},
  year={2022},
  publisher={Elsevier}
}

@article{cong2023scaling,
  title={Scaling smart contracts via layer-2 technologies: Some empirical evidence},
  author={Cong, Lin William and Hui, Xiang and Tucker, Catherine and Zhou, Luofeng},
  journal={Management Science},
  volume={69},
  number={12},
  pages={7306--7316},
  year={2023},
  publisher={INFORMS}
}

@inproceedings{shani2020adaptive,
  title={Adaptive trust region policy optimization: Global convergence and faster rates for regularized mdps},
  author={Shani, Lior and Efroni, Yonathan and Mannor, Shie},
  booktitle={Proceedings of the AAAI Conference on Artificial Intelligence},
  volume={34},
  number={04},
  pages={5668--5675},
  year={2020}
}

@inproceedings{confIJCAI0001HH0Z20,
  author       = {Zhuang Liu and
                  Degen Huang and
                  Kaiyu Huang},
  title        = {FinBERT: {A} Pre-trained Financial Language Representation Model for Financial Text Mining},
  booktitle    = {{IJCAI} 2020},
  pages        = {4513--4519},
  year         = {2020},
  url          = {https://doi.org/10.24963/ijcai.2020/622},
  doi          = {10.24963/IJCAI.2020/622}
}

@inproceedings{iclrBambergerB0B25,
  author       = {Jacob Bamberger and
                  Federico Barbero and
                  Xiaowen Dong and
                  Michael M. Bronstein},
  title        = {Bundle Neural Network for message diffusion on graphs},
  booktitle    = {The Thirteenth International Conference on Learning Representations,
                  {ICLR} 2025, Singapore, April 24-28, 2025},
  publisher    = {OpenReview.net},
  year         = {2025},
  url          = {https://openreview.net/forum?id=scI9307PLG}
}

@article{istrSaidaneTSG25,
  author       = {Samia Saidane and
                  Francesco Telch and
                  Kussai Shahin and
                  Fabrizio Granelli},
  title        = {Deep GraphSAGE enhancements for intrusion detection: Analyzing attention
                  mechanisms and {GCN} integration},
  journal      = {Journal of Information Security and Applications},
  volume       = {90},
  pages        = {104013},
  year         = {2025},
  url          = {https://doi.org/10.1016/j.jisa.2025.104013},
  doi          = {10.1016/J.JISA.2025.104013}
}

@article{hochreiter1997long,
  author       = {Sepp Hochreiter and
                  J{\"{u}}rgen Schmidhuber},
  title        = {Long Short-Term Memory},
  journal      = {Neural Computation},
  volume       = {9},
  number       = {8},
  pages        = {1735--1780},
  year         = {1997},
  url          = {https://doi.org/10.1162/neco.1997.9.8.1735},
  doi          = {10.1162/NECO.1997.9.8.1735}
}

@inproceedings{pennington2014glove,
  author       = {Jeffrey Pennington and
                  Richard Socher and
                  Christopher D. Manning},
  title        = {Glove: Global Vectors for Word Representation},
  booktitle    = {Proceedings of the 2014 Conference on Empirical Methods in Natural
                  Language Processing, {EMNLP} 2014, October 25-29, 2014, Doha, Qatar,
                  {A} meeting of SIGDAT, a Special Interest Group of the {ACL}},
  pages        = {1532--1543},
  publisher    = {{ACL}},
  year         = {2014},
  url          = {https://doi.org/10.3115/v1/d14-1162},
  doi          = {10.3115/V1/D14-1162}
}

@inproceedings{kipf2016semi,
  author       = {Thomas N. Kipf and
                  Max Welling},
  title        = {Semi-Supervised Classification with Graph Convolutional Networks},
  booktitle    = {5th International Conference on Learning Representations, {ICLR} 2017,
                  Toulon, France, April 24-26, 2017, Conference Track Proceedings},
  publisher    = {OpenReview.net},
  year         = {2017},
  url          = {https://openreview.net/forum?id=SJU4ayYgl}
}

@article{deepseek2024,
  author       = {DeepSeek{-}AI and
                  Daya Guo and
                  Dejian Yang and
                  Haowei Zhang and
                  Junxiao Song and
                  Ruoyu Zhang and
                  Runxin Xu and
                  Qihao Zhu and
                  Shirong Ma and
                  Peiyi Wang and
                  Xiao Bi and
                  Xiaokang Zhang and
                  Xingkai Yu and
                  Yu Wu and
                  Z. F. Wu and
                  Zhibin Gou and
                  Zhihong Shao and
                  Zhuoshu Li and
                  Ziyi Gao and
                  Aixin Liu and
                  Bing Xue and
                  Bingxuan Wang and
                  Bochao Wu and
                  Bei Feng and
                  Chengda Lu and
                  Chenggang Zhao and
                  Chengqi Deng and
                  Chenyu Zhang and
                  Chong Ruan and
                  Damai Dai and
                  Deli Chen and
                  Dongjie Ji and
                  Erhang Li and
                  Huajian Xin and
                  Huazuo Gao and
                  Hui Qu and
                  Hui Li and
                  Jianzhong Guo and
                  Jiashi Li and
                  Jiawei Wang and
                  Jingchang Chen and
                  Jingyang Yuan and
                  Junjie Qiu and
                  Junlong Li and
                  J. L. Cai and
                  Ruiqi Ge and
                  Ruisong Zhang and
                  Ruizhe Pan and
                  Runji Wang and
                  R. J. Chen and
                  R. L. Jin and
                  Ruyi Chen and
                  Shanghao Lu and
                  Shangyan Zhou and
                  Shanhuang Chen and
                  Shengfeng Ye and
                  Shiyu Wang and
                  Shuiping Yu and
                  Shunfeng Zhou and
                  Shuting Pan and
                  S. S. Li},
  title        = {DeepSeek-R1: Incentivizing Reasoning Capability in LLMs via Reinforcement
                  Learning},
  journal      = {CoRR},
  volume       = {abs/2501.12948},
  year         = {2025},
  url          = {https://doi.org/10.48550/arXiv.2501.12948},
  doi          = {10.48550/ARXIV.2501.12948},
  eprinttype    = {arXiv},
  eprint       = {2501.12948}
}

@article{sims2003implications,
title={Implications of rational inattention},
author={Sims, Christopher A},
journal={Journal of Monetary Economics},
volume={50},
number={3},
pages={665--690},
year={2003}
}

@article{mackowiak2023rational,
  title={Rational inattention: A review},
  author={Ma{\'c}kowiak, Bartosz and Mat{\v{e}}jka, Filip and Wiederholt, Mirko},
  journal={Journal of Economic Literature},
  volume={61},
  number={1},
  pages={226--273},
  year={2023}
}

@article{gabaix2019behavioral,
  title={Behavioral inattention},
  author={Gabaix, Xavier},
  booktitle={Handbook of behavioral economics: Applications and foundations 1},
  volume={2},
  pages={261--343},
  year={2019},
  publisher={Elsevier}
}

@article{wager2018estimation,
  author       = {Hugo Bodory and
                  Federica Mascolo and
                  Michael Lechner},
  title        = {Enabling Decision Making with the Modified Causal Forest: Policy Trees
                  for Treatment Assignment},
  journal      = {Algorithms},
  volume       = {17},
  number       = {7},
  pages        = {318},
  year         = {2024},
  url          = {https://doi.org/10.3390/a17070318},
  doi          = {10.3390/A17070318}
}

@article{park2023conceptual,
  title={The conceptual flaws of decentralized automated market making},
  author={Park, Andreas},
  journal={Management Science},
  volume={69},
  number={11},
  pages={6731--6751},
  year={2023},
  publisher={INFORMS}
}

@article{pearl2009causality,
  title={Causal inference in statistics: An overview},
  author={Pearl and Judea},
  journal={Statistics Surveys},
  volume={3},
  pages={96-146},
  year={2009},
}

@article{chernozhukov2018double,
  title={Double/debiased machine learning for treatment and structural parameters},
  author={ Chernozhukov, Victor  and  Chetverikov, Denis  and  Demirer, Mert  and  Duflo, Esther  and  Hansen, Christian  and  Newey, Whitney K.  and  Robins, James },
  journal={The Econometrics Journal},
  volume={21},
  pages={C1–C68},
  year={2018},
}

%%%%%%%%%%%%%%%%%
\end{document}